%% file: neurips_2025.tex
\documentclass{article}

\usepackage[square,numbers,sort&compress]{natbib}

\usepackage[preprint]{neurips_2025}

\usepackage[utf8]{inputenc} 
\usepackage[T1]{fontenc}    
\usepackage{hyperref}       
\usepackage{url}            
\usepackage{booktabs}       
\usepackage{amsfonts}       
\usepackage{nicefrac}       
\usepackage{microtype}      
\usepackage{xcolor}         

\usepackage{makecell}      
\usepackage{adjustbox}
\usepackage{multirow}
\usepackage{booktabs}
\usepackage{tabularx} 
\usepackage{placeins}
\usepackage{graphicx}
\usepackage{amsmath}
\usepackage{wrapfig}
\usepackage{caption}
\usepackage{subcaption}

\def\Snospace~{\S{}}

\title{LED: LLM Enhanced Open-Vocabulary Object Detection without Human Curated Data Generation}

\author{
Yang Zhou\quad
Shiyu Zhao\quad
Yuxiao Chen\quad
Zhenting Wang\quad
Can Jin\quad
Dimitris N.\ Metaxas\\[3pt]
Rutgers University\\
\texttt{\{eta.yang,\,sz553,\,yc984,\,zhenting.wang,\,can.jin,\,dnm\}@rutgers.edu}
}

\newcommand{\LED}{\protect\scalebox{1.2}{L}\protect\scalebox{1.0}{ED}}

\begin{document}

\maketitle

\input{sec/0_abstract}    
\input{sec/1_intro_v2}
\input{sec/2_relate}
\input{sec/3_approach}

\input{sec/4_experiments}

\input{sec/5_conclusion}

\bibliography{neurips_2025}


\appendix

\input{sec/X_suppl}



\end{document}

%% file: sec/0_abstract.tex
\begin{abstract}
Large foundation models trained on large-scale vision–language data can boost Open-Vocabulary Object Detection (OVD) via synthetic training data, yet the hand-crafted pipelines often introduce bias and overfit to specific prompts.
We sidestep this issue by \emph{directly fusing} hidden states from Large Language Models (LLMs) into detectors—an avenue surprisingly under-explored.
This paper presents a systematic method to enhance visual grounding by utilizing decoder layers of the LLM of a MLLM. 
We introduce a zero-initialized cross-attention adapter to enable efficient \textbf{knowledge-fusion} from LLMs to object detectors, an new approach called \textit{\LED} (\textbf{L}LM \textbf{E}nhanced Open-Vocabulary Object \textbf{D}etection).
We find that intermediate LLM layers already encode rich spatial semantics; adapting only the early layers yields most of the gain.
With Swin-T as the vision encoder, Qwen2-0.5B +\LED{} lifts GroundingDINO by \textbf{3.82 \%} on OmniLabel at just \textbf{8.7 \%} extra GFLOPs, and a larger vision backbone pushes the improvement to \textbf{6.22 \%}.
Extensive ablations on adapter variants, LLM scales and fusion depths further corroborate our design. 

\end{abstract}

%% file: sec/1_intro_v2.tex
\section{Introduction}
\label{sec:intro_v2}

\begin{wrapfigure}[28]{R}{0.5\textwidth}
    \centering
    \includegraphics[width=1\linewidth]{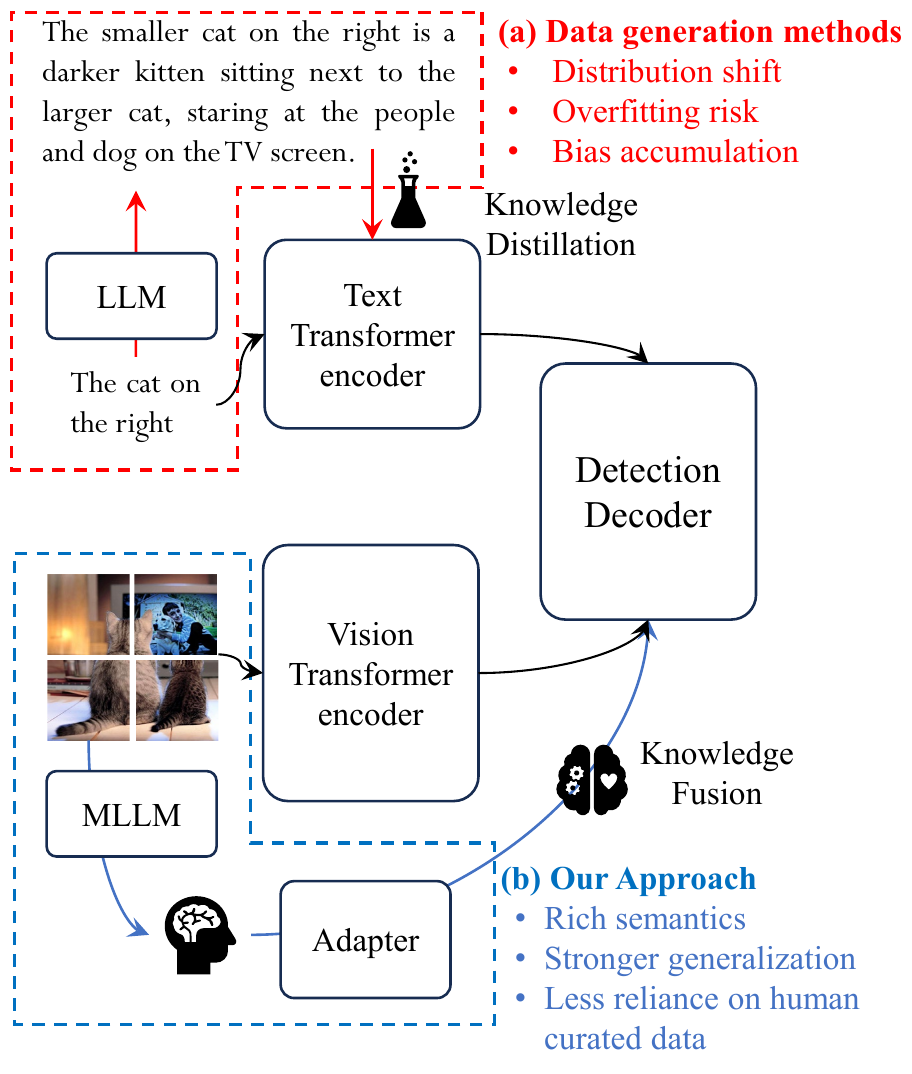}
\caption{Overview of open vocabulary object detection approaches. (a) Data generation methods prone to distribution shift and overfitting; (b) Our proposed adapter-based framework directly leverages knowledge from the MLLM to the detection decoder.}
\label{approachs_compare}
\vspace{-8pt}
\end{wrapfigure}

Open Vocabulary object Detection (OVD) localizes objects described by free-form text, significantly increasing the size of the label space and enabling broader applications, compared to the traditional closed-set object detection with a fixed label space.
Significant prior efforts~\cite{zhao2022exploiting, minderer2023scaling,li2023desco, yao2024detclipv3, zhao2024taming} focus on exploring large foundation models trained on image and/or text data to generate high-quality data for OVD, greatly boosting performance on challenge benchmarks~\cite{gupta2019lvis,schulter2023omnilabel,xie2023described}.
Their success demonstrates that large foundation models implicitly capture rich semantic knowledge and reasoning capabilities through pretraining, which are beneficial for accurately localizing objects under complex, free-form textual queries.

However, existing methods rely on hand-crafted rules to query large foundation models for synthetic data. 
For example, VL-PLM~\citep{zhao2022exploiting} and OWLv2~\cite{minderer2023scaling} design a delicate procedure with heuristics to use pretrained vision and language models (VLMs) to generate pseudo labels for unseen categories that are not included in the human provided data. DesCo~\cite{li2023desco} and GenEnhancedNegs~\cite{zhao2024generating} assume that detailed or negative descriptions are important training signals for OVD, and thus leverage large language models (LLMs) to create those descriptions.
Those methods are well designed for certain aspects (e.g. unseen categories and descriptions) to improve OVD models, \textbf{which are biased and cannot fully leverage knowledge gained during the pretraining of large foundation models}.
 
We argue that the \emph{latent representations} of large multimodal language models already encode rich visual–linguistic concepts that need not be materialised into explicit labels. Building on this observation, we introduce \LED, an \textbf{L}LM \textbf{E}nhanced Open-Vocabulary Object \textbf{D}etection that realises \textbf{a novel knowledge-fusion paradigm}: instead of learning a detector \emph{from scratch} or augmenting training with generated data, we \emph{directly fuse} intermediate hidden states from a frozen MLLM into popular mainstream detector via a lightweight adapter.
Specifically, as shown in \autoref{approachs_compare}(a), existing approaches primarily distill LLM knowledge into the object detector via explicit data generation—such as synthesizing negative examples to mitigate hallucinations \cite{you2023ferret,zhai2023halle, rohrbach2018object, yin2024woodpecker, zhao2024generating}, or creating detailed annotations of spatial and logical relationships \cite{li2023desco}. However, these explicit generation methods inherently constrain model robustness and generalizability. In contrast, as illustrated in \autoref{approachs_compare}(b), our method \textbf{directly fuses the high-dimensional semantic knowledge} encapsulated within LLM hidden states into the detector, thereby preserving richer pretrained knowledge without explicit data augmentation.

We apply \LED~to popular open-source object detectors.
GroundingDino~\cite{liu2024grounding}. 
Our method significantly improves performance on prevailing benchmarks, including RefCOCO~\cite{yu2016modeling} and OmniLabel benchmark~\cite{schulter2023omnilabel}. Besides, we find that hidden states from the first several layers of a MLLM are rich enough to improve the OVD models. Probably, \textbf{MLLMs focuses more on vision tokens in early layers and then on text tokens in later layers}, as illustrated in \autoref{Methodology}. 
Thus, vision tokens of deeper layers are not quite updated and embody similar knowledge as earlier ones.

Our contributions are summarized as:

(1) We introduce a lightweight adapter that directly \textbf{fuses} frozen MLLM hidden representations into any object detector, establishing a unified knowledge‐injection pipeline and eliminating the need for expensive data synthesis.

(2) We systematically explore various fusion designs—layer selection, injection modality, adapter structure, etc.—and identify the most effective pipeline, offering clear guidelines for future work.

(3) We conduct extensive experiments, demonstrating that our approach not only improves OVD models across various benchmarks, but also achieves benefits comparable or even better to those obtained through data generation.




%% file: sec/2_relate.tex
\section{Relate Work}
\label{sec:formatting}
{\bf Multi-Modal Large Language Models: }
Recent MLLMs have driven visual–language integration through modality alignment \cite{jin2025visual, jin2025lorvp}. CLIP pioneered contrastive learning on large-scale image–text pairs \cite{radford2021learning}, and BLIP enhanced alignment with curated data \cite{li2022blip,li2023blip}. LLaVA applies GPT-4–generated multimodal instruction tuning to boost reasoning \cite{liu2024visual}. InternVL’s progressive alignment scales vision–language models to SOTA in foundation models \cite{chen2024expanding,chen2024far,chen2024internvl,gao2024mini,wang2024mpo}. LLaMA Adapter uses a zero-initialized attention mechanism with just 1.2 M parameters for prompt-based tuning, matching full fine-tuning \cite{zhang2023llama}. \LED~instead fuses high-level MLLM embeddings directly into downstream detectors.

\noindent
{\bf Open-Set Object Detection:}
While MLLMs excel at integrating textual and visual modalities, open-world detection remains an unsolved challenge that calls for more flexible category representations. DINO advances closed-set detection through contrastive de-noising in DETR frameworks \cite{zhang2022dino, carion2020end}, its reliance on fixed category definitions limits open-world applicability. This motivates open-set detection methods that leverage language generalization: OV-DETR employs CLIP embeddings as cross-modal queries for category-agnostic box decoding \cite{zareian2021open}, and GLIP reformulates detection as visual grounding with phrase-region alignment \cite{li2022grounded, gao2024clip}. GroundingDINO subsequently refines these grounding mechanisms, establishing new benchmarks in zero-shot detection \cite{liu2024grounding}.

\noindent
{\bf Vision-Language Grounding:}
With open-set capabilities taking shape, further progress demands robust vision-language grounding that handles diverse and unconstrained data distributions. Recent advances in leveraging vision-language models (VLMs) for open-vocabulary and open-ended object detection predominantly rely on large-scale synthetic or generated datasets. Techniques such as prompt-driven visual-text adaptation \cite{yao2022detclip,long2023fine} and generative augmentation of negative samples \cite{zhao2024generating} have significantly improved model performance by enriching training data with diversified visual and textual examples. Similarly, generative region-language pretraining frameworks \cite{lin2024generative} and unified object-centric grounding methods \cite{ren2024dino} also heavily rely on generated or automatically curated datasets, demonstrating their efficiency and scalability.

However, synthetic data often misaligns with real-world distributions, introducing biases that degrade performance under natural variations and unseen scenarios. Moreover, overreliance on generated samples can lead to overfitting to specific dataset artifacts, limiting adaptability in truly open-world settings. While approaches like GLIP \cite{li2022grounded} and DetCLIP \cite{yao2022detclip} improve grounding via enriched annotations, their dependence on synthetic constructs remains a core limitation.


%% file: sec/3_approach.tex
\section{Methodology}
\label{Methodology}

\begin{wrapfigure}{R}{0.5\textwidth}
    \centering
    \includegraphics[width=1\linewidth]{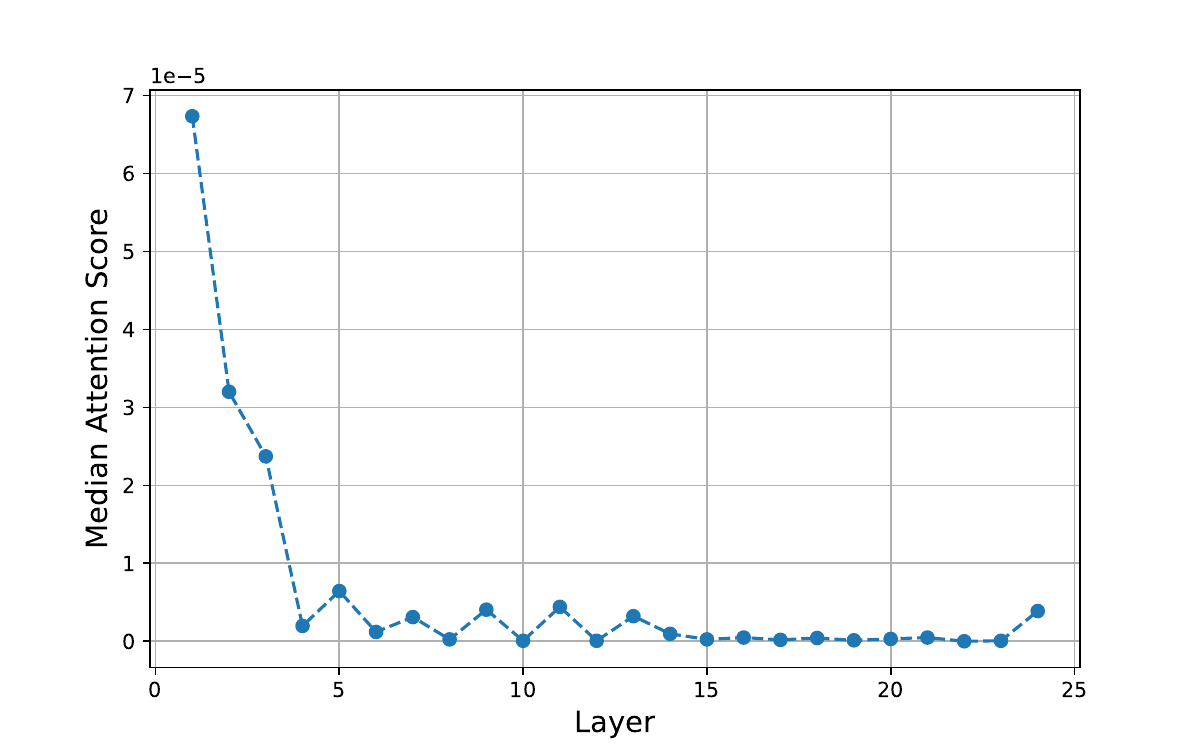}
    \caption{Median distribution of attention scores across different layers in Qwen2-0.5B decoder. }
\label{median_attention_scores}
\vspace{-8pt}
\end{wrapfigure}

Our framework establishes an efficient pipeline for transferring multimodal knowledge from LLMs to OVD models via intermediate feature adaptation. As shown in \autoref{median_attention_scores},  we analyze token attention changes across different layers of the LLM decoder for Qwen2-0.5B. In the early layers of the LLMs decoder (e.g., layers 0 to 4), the median attention score decreases significantly and fluctuates slightly in the subsequent 20 layers. This indicates that the \textbf{MLLMs already exhibit strong attention or encoding capabilities for input image tokens in the early layers}, while the later layers may focus more on fine-tuning task- or semantics-related text tokens \autoref{Attention Maps in Qwen2-0.5B Decoding} (in the supplement).

\begin{figure}
    \centering
    \includegraphics[width=1\linewidth]{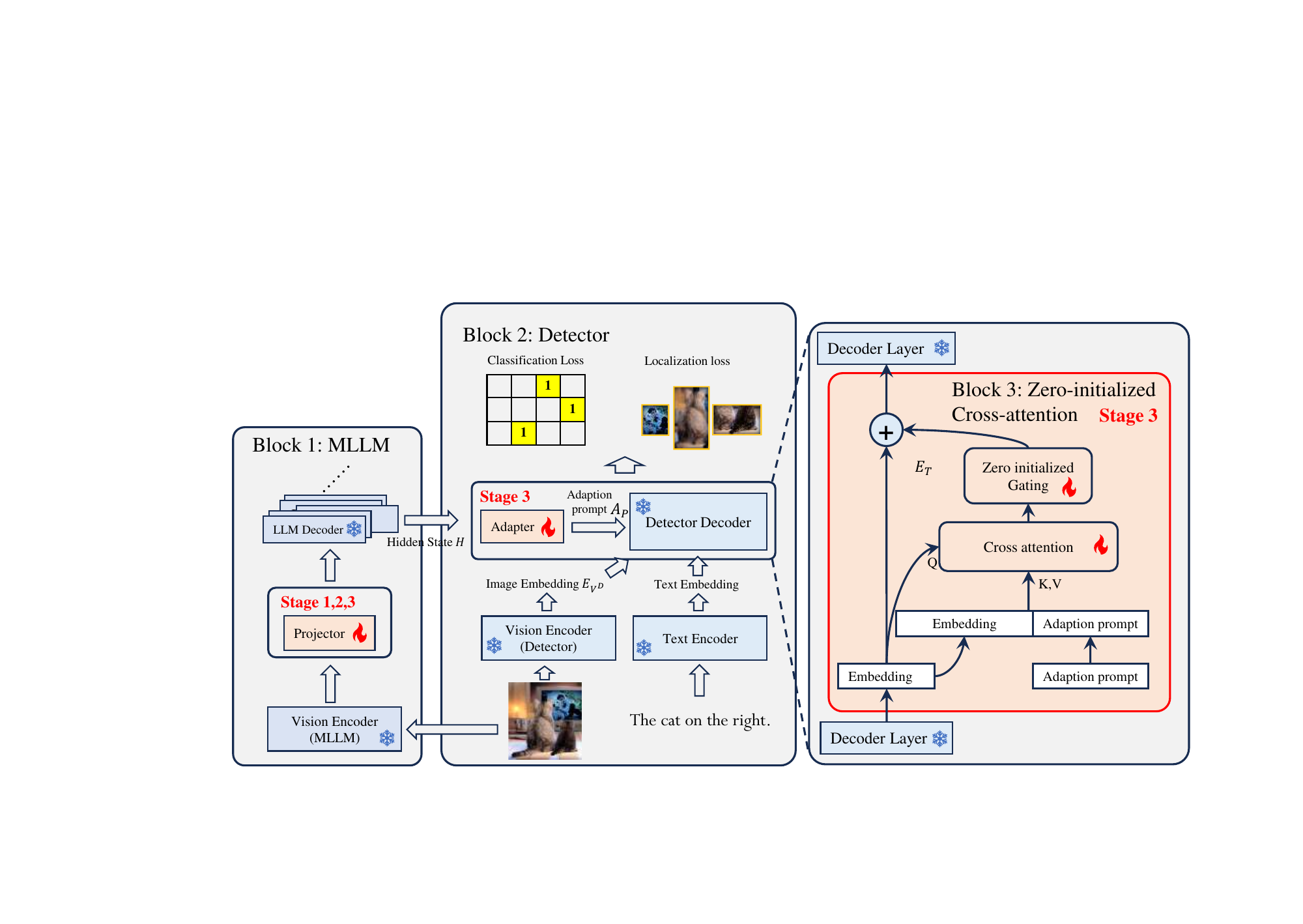}
    \caption{The architecture of \LED. We present the overall architecture, MLLM, detector, zero-initialized cross-attention in Block 1, 2, 3, respectively. In stage 1 and 2, a trained projector is used to align the frozen Vision Encoder with the LLM. In stage 3, a zero-initialized adapter is trained to guide the detector with adaptation prompts, while the projector still participate training to adapt to downstream tasks.}
    \label{architecture}
     \vspace{-8pt}
\end{figure}

To validate this concept, we design an architecture illustrated in \autoref{architecture}. The architecture comprises four key components: (1) an efficient cross-modal alignment strategy in \autoref{Cross-modal Alignment}; (2) an MLLM for semantic adaptation prompts generation in \autoref{Semantic Prompt Generation}; (3) a universal end-to-end detector in \autoref{Grounding Detector}; (4) a zero-initialized cross-attention adapter for feature fusion \autoref{Zero-initialized Cross-attention}.

\subsection{Cross-modal Alignment}
\label{Cross-modal Alignment}
\LED~aims to provide a comprehensive solution by leveraging the extensive knowledge of LLMs to enhance semantic image understanding. In general, MLLMs for text and images require a vision encoder to generate visual tokens for LLMs. Thus, cross-modal alignment is crucial for ensuring that LLMs can interpret the visual tokens generated by the vision encoder and assign semantic meaning to them. To facilitate effective knowledge transfer while maintaining computational efficiency, we establish a shared vision encoder between the MLLM and the detector through a three-stage alignment process:

\textbf{Stage 1 - Pretraining:} We employ a vision encoder and LLM to pretrain the MLP projection layer to align feature spaces.

\textbf{Stage 2 - Finetuning:} We fine-tune the model on a specific task dataset to align the vision encoder and LLM at a fine-grained level and improve instruction-following capabilities.

\textbf{Stage 3 - Adapter Training:} With the MLLM and detector frozen, we train the grounding decoder adapter (Block 3) and the MLP layer in the MLLM to integrate semantic cues from its intermediate features, thereby enabling the LLM to better focus on grounding tasks when processing visual tokens.

\subsection{Semantic Prompt Generation}
\label{Semantic Prompt Generation}
As illustrated in \autoref{architecture} Block 1, we employ an MLLM that processes the same input images shared with the detector. The MLLM integrates visual features with system prompts and provides the resulting embeddings as input to the LLM.

We extract intermediate representations from the $\ell_{LM}$-th decoder layer of an $n$-layer language model, referred to as the hidden state embedding $\mathbf{H}$. After obtaining the $\mathbf{H}$, the vision embedding $\mathbf{E_V}^L$ and text embeddings $\mathbf{E_T}$ are derived by truncating the embeddings based on the positions of the image and text tokens:

\begin{equation}
\begin{aligned}
&\mathbf{H}_{\ell_{LM}} = \mathrm{Decoder}_{\ell_{LM}}(\mathbf{H}_{\ell_{LM}-1}), \ell_{LM} \in \{1,...,n\}\\
&\mathbf{E_V}^L, \mathbf{E_T} = \text{Truncate}(\mathbf{H}_{\ell_{LM}})
\end{aligned}
\end{equation}

\subsection{Grounding Detector}
\label{Grounding Detector}
An End-to-End Transformer-based Detector architecture, as shown in \autoref{architecture} Block 2. For every (Image, Text) pair, this Block employs a vision encoder and a text encoder to independently extract visual and textual features. These features are then passed into a cross-modality decoder, which interacts with both modalities to refine and update the features. The final output queries generated by the last layer of the decoder are used to predict bounding boxes for objects and extract their corresponding textual phrases \cite{liu2024grounding, zhang2022dino, carion2020end}.

\subsection{Zero-initialized Cross-attention}
\label{Zero-initialized Cross-attention}

An adapter design (Block 3 in \autoref{architecture}) effectively refines the vision embedding $\mathbf{E_V}^L$ and text embedding $\mathbf{E_T}$ from the LLM to generate adaptation prompts with detector features through zero-initialized cross-attention. We apply a convolutional layer to process the MLLM's vision embeddings $\mathbf{E_V}^L$ to get adaptation prompts $\mathbf{A_P}$.
\begin{equation}
\mathbf{A_P} = \mathrm{Conv2D}(\mathbf{E_V}^L)  \in \mathbb{R}^{B \times L \times d}
\end{equation}
where $B$ is batch size, $L$ is the length of adaptation prompts, decomposed into the height $h$ and width $w$ of the Vision embedding $\mathbf{E_V}^L$ when performing convolution processing.

A zero-initialized cross-modal attention mechanism is designed based on a small gating unit. It is worth noting that since the weights of the entire adapter are randomly initialized, directly providing the adaptation prompt to the detector decoder introduces disordered noise, which is detrimental to the training of the network. Given a decoder embedding $\mathbf{E_D}^{\ell_D} \in \mathbb{R}^{B \times T \times d}$ from $\ell_{D}$th decoder layer and adaptation prompts $\mathbf{A_P} \in \mathbb{R}^{B \times L \times d}$, our zero-initialized cross-modal attention mechanism operates as follows:

\begin{equation}
\begin{aligned}
\mathbf{Q} &= \mathbf{E_D}\mathbf{W}_Q \in \mathbb{R}^{B \times T \times h \times d_h} \\
\mathbf{K} &= \mathcal{C}(\mathbf{A_P}\cdot\mathbf{W}_K, \mathbf{E_D}\mathbf{W}_K) \in \mathbb{R}^{B \times (L+T) \times h \times d_h} \\
\mathbf{V} &= \mathcal{C}(\mathbf{A_P}\cdot\mathbf{W}_V, \mathbf{E_D}\mathbf{W}_V) \in \mathbb{R}^{B \times (L+T) \times h \times d_h}
\end{aligned}
\end{equation}

where $\mathcal{C}(\cdot)$ denotes concatenation along the sequence dimension, $h$ is the number of attention heads, and $d_h = d/h$ represents the dimension per head.
$\mathbf{W}_Q, \mathbf{W}_K, \mathbf{W}_V$ are the linear projection matrices used in the attention mechanism to map to $Q, K, V$.

\begin{equation}
\begin{aligned}
\mathbf{S} = &\frac{\tilde{\mathbf{Q}}\tilde{\mathbf{K}}^\top}{\sqrt{d_h}} + \mathbf{M} \in \mathbb{R}^{B \times h \times T \times (L+T)} \\
\tilde{\mathbf{S}} = & \mathcal{C}(\tanh(\mathbf{g}) \odot \text{softmax}(\mathbf{S}_{:,0:L} ), \text{softmax}(\mathbf{S_{:,L+1:L+T}})) \in \mathbb{R}^{B \times h \times T \times (L+T)}
\end{aligned}
\end{equation}

where $\tilde{\mathbf{Q}}, \tilde{\mathbf{K}}$ is defined as $\text{RoPE}(\mathbf{Q}, \mathbf{K}; \Theta)$ with angular parameter matrix $\Theta$; $\mathbf{S}$ represents the raw attention score matrix, and $\tilde{\mathbf{S}}$ is attention weight matrix obtained by applying $\text{softmax}$; $\mathbf{g} \in \mathbb{R}^{1 \times h \times 1 \times 1}$ is a learnable gating vector initialized to $\mathbf{0}$. and an activation function $\tanh(\mathbf{\cdot})$ is adopted to regulate the scale of $\mathbf{g}$ to into $-1\sim1$. If adaptation prompts are initialized randomly, they may introduce noise or interfere with word tokens during the initial stages of training, thereby compromising the stability and effectiveness of the fine-tuning process.

Finally a linear projection layer calculate the output as adaptation prompts and add on decoder embedding from $\ell_D$th decoder layer as:
\begin{equation}
\mathbf{E_D}^{\ell_D} =\mathbf{E_D}^{\ell_D-1} + \mathbf{Linear}(\tilde{\mathbf{S}}\mathbf{V}) \in \mathbb{R}^{B \times T \times d}
\end{equation}

%% file: sec/4_experiments.tex
\section{Experiments}
In this section, we first detail the datasets and implementation specifics \autoref{Dataset and Implementation Details}). Next, through evaluations on the OmniLabel benchmark\cite{schulter2023omnilabel}, we analyze strategies for sharing vision decoder features (see \autoref{Vision Decoder Share from Detector}), investigate the impact of different adapter architectures (see \autoref{adapter_architecture_Analysis}), and observe performance improvements across various scales of LLMs for visual detection and grounding tasks (see \autoref{Different LLMs}). Finally, we conduct ablations to precisely quantify the contributions of MLLM decoder hidden states at various layers to detection performance (see \autoref{Ablations}).

\subsection{Dataset and Implementation Details}
\label{Dataset and Implementation Details}
\noindent
{\bf MLLMs alignment:}
The dataset used to train the projector follows the approach outlined in the ShareGPT-4V plan \cite{chen2023sharegpt4v}. Our architecture adopts a multi-stage training paradigm, incorporating a carefully designed data composition. For \emph{InternVL2-1B}, we employ a \textbf{two-phase strategy}:
\begin{enumerate}
    \item[] \textit{Stage 1}: We leverages \texttt{LAION-CC-SBU} (558K samples) \cite{liu2023visual} to establish cross-modal alignment through diverse visual-linguistic interactions.
    \item[] \textit{Stage 2}: We combine \texttt{ShareGPT4V} (102K) and \texttt{SFT} (665K) \cite{chen2023sharegpt4v} for instruction tuning, augmented with domain-specific datasets \autoref{sec:Dataset Introduction} (see Supplement) to enhance structured visual understanding.
\end{enumerate}

\noindent
{\bf Stage 3 - Adapter training:}
We employ Grounding DINO \cite{liu2024grounding} pretrain on Object365, GoldG and Cap4M as a unified detection pre-training framework across three datasets: \texttt{Objects365} \cite{shao2019objects365}, \texttt{COCO2017}, and \texttt{Flickr30k} \cite{young2014image}. We implement our framework using InternVL2-1B (Qwen2-0.5B backbone) as the MLLM \cite{chen2024expanding, wang2024mpo, gao2024mini, chen2024far, chen2024internvl} and Open-GroundingDINO \cite{liu2024grounding} as the detection codebase. All experiment configurations as shown in \autoref{sec:Experiment Configuration} (see Supplement)


\subsection{Evaluation}
We evaluated and analyzed adapters under different architectures based on the OmniLabel benchmark \cite{schulter2023omnilabel}. Additionally, we compared the performance results of the hidden state acquisition scheme in InternVL2 with GroundingDINO (abbreviated as G-DINO in tables).

\subsubsection{Vision Decoder Share from Detector}
\label{Vision Decoder Share from Detector}

The rich semantic understanding of the MLLM primarily originates from the LLM rather than the vision encoder. Thus, we adopt GroundingDINO's vision encoder architecture and weights to provide visual inputs for both the detector and MLLM.
We conduct MLLM training in two stages (see \autoref{architecture}): pretraining (stage 1) and finetuning (stage 2), freezing both the vision encoder and LLM during training. \autoref{MLLM Alignment (Swin-T & Qwen2-0.5B)} (supplementary) provides MME evaluations of alignment progress between stages.

Specifically, the shared Swin-Tiny vision encoder output undergoes a pixel shuffle operation (scaling factor 4), reducing spatial dimensions $(h,w)$ by a factor of 4 while expanding the channels by a factor of 16. This transformed output is then truncated to match the MLP projection layer dimension (4096), aligning the feature spaces through pretrained MLP layers:

\begin{equation}
    \hat{f}_v = \text{MLP}(\text{Truncate}(\text{Repeat}(\text{PixelShuffle}_{4}(\text{Vision Encoder}(I)))))
\end{equation}

where $I$ denotes the input image and $\hat{f}_v$ represents the visual features projected. $\text{PixelShuffle}_{4}(\cdot)$ denotes the pixel shuffle operation with a scaling factor of 4, $\text{Repeat}(\cdot)$ represents the repetition of channels, $\text{Truncate}(\cdot)$ ensures that the channel dimension matches the MLP's input requirements.





\begin{figure}[!tb]
  \centering
  \begin{minipage}[t]{0.52\linewidth}
    \vspace{0pt}
    \captionof{table}{Model params, compute cost, and inference latency; most overhead stems from the LLM’s two layers, with minimal adapter cost.}
    \label{params and flops}
    \centering
    \renewcommand{\arraystretch}{1.1}
    \begin{tabular}{p{0.35\linewidth} p{0.12\linewidth} p{0.14\linewidth} p{0.18\linewidth}}
      \toprule
      \textbf{Framework} & \textbf{Params} & \textbf{GFLOPs} & \textbf{Latency}\\
      \midrule
      G-DINO & 172M & 412G & 299.3 ms \\
      \textbf{\LED} & +\textbf{58M} & +\textbf{36G} & +\textbf{52.3ms} \\
      \hspace{0.5em}-Adapter   & +2.2M & +2.0G & +2.0ms\\
      \hspace{0.5em}-Qwen2-0.5B $2_L$ & +56M & +34G & +50.5ms\\

      \bottomrule
    \end{tabular}
  \end{minipage}\hfill
  \begin{minipage}[t]{0.46\linewidth}
    \vspace{0pt}
    \captionof{table}{OmniLabel evaluation for the three-stage pipeline and shared vision decoder, relative to GroundingDINO.}
    \label{omnilabel_evaluation_MLLMs_ablation}
    \centering
    \begin{tabular}{p{0.30\linewidth} p{0.18\linewidth} p{0.14\linewidth} p{0.14\linewidth}}
      \toprule
      \multirow{2}{*}{\textbf{Framework}} & \textbf{AP} & \textbf{AP} & \textbf{AP} \\
       & descr,categ & categ & descr \\
      \midrule
      G-DINO & 21.69\% & 33.25\% & 16.09\% \\
      \makecell[l]{\textbf{\LED} \\ \hspace{0.5em}-Qwen2-0.5B}
        & \makecell[l]{25.51\%\\\textcolor{blue}{$\uparrow$3.82\%}}
        & \makecell[l]{33.01\%\\\textcolor{blue}{$\uparrow$0.25\%}}
        & \makecell[l]{20.28\%\\\textcolor{blue}{$\uparrow$4.19\%}} \\
      \bottomrule
    \end{tabular}
  \end{minipage}
\end{figure}

Finally, when training the adapter (stage 3 in \autoref{architecture}), both the LLM and the MLP layers are trained simultaneously to adapt to downstream tasks. Image embeddings from Swin-Tiny (used in GroundingDINO) are fed into the LLM after alignment by a projector. The pretrained weights for Swin-Tiny, which are the same as those used in GroundingDINO and provided by Open-GroundingDINO \cite{OpenGroundingDino}, are utilized and frozen during every stage. 

We measured the GFLOPs for each architecture and selected the $2_{nd}$ decoder layer, removing subsequent layers. As shown in Table \ref{params and flops}, \LED~requires only 8.7\% more FLOPs than GroundingDINO, with the majority of the computational cost attributed to the LLM's decoder. 

The OmniLabel benchmark \cite{schulter2023omnilabel} for our method is presented in Table \ref{omnilabel_evaluation_MLLMs_ablation} with an improvement of $3.82\%$ over GroundingDINO. This gain stems from the adaptation prompts generated by the LLM.

\subsubsection{Adapter Architecture Analysis}

In MLLMs, how the vision encoder aligns with the LLM decisively affects image-token understanding. We adopt the InternVL ViT backbone (pre-trained) for InternVL2-1B/2B/8B and compare four adapter designs that extract \emph{adaptation prompts} from decoder hidden states while coping with truncated text embeddings.

\textbf{Arch.~I — Double Cross-Modal Fusion.} As shown in the \autoref{adapter_architecture_1} (in the supplement), owing to the robust cross-modal feature fusion capabilities of MLLMs, the text embeddings $\mathbf{E_T}$ extracted from the hidden state, encompass substantial information relevant to image understanding. Furthermore, these embeddings can guide the Detector by leveraging user-provided instructions. The LLM input is formulated as a structured prompt that integrates task instructions and object captions, adhering to GroundingDINO's annotation protocol:
    
    \begin{quote}
    \textit{`$<$user$>$: [Instruction: 'Detect objects in this image. If present, locate them with bounding boxes.'] + [Caption: 'The cat on the right']}
    \end{quote}
    
    We extract image/text embeddings $\mathbf{E_T}/\mathbf{E}_{V^D}$ from the decoder's hidden states using positional segmentation. The cross-attention mechanism operates through two stages: (1) \textit{Text-guided image fusion}: Image embeddings as queries attend to text embeddings (keys/values); (2) \textit{Prompt-enhanced detection}: Detector queries attend to fused embeddings (keys/values)

    Through two cross-attention mechanisms, we integrate the Adaptation prompt, which combines images and text, back into the Image embedding of the Detector through a zero-initialized gating:
    
    \begin{equation}
        \mathbf{A_P} = Gating(CrossAtt(CrossAtt(\mathbf{E_V^L},\mathbf{E_T}),\mathbf{E}_{V^D}))
    \end{equation}
    
\textbf{Arch.~II — Late Prompt Projection.} 
    We drop the second cross-modal block.  
    As shown in the \autoref{adapter_architecture_23} (see Supplement),  a $3{\times}3$\,/2 convolution maps the fused prompt to the detector space and injects it into the last decoder layer ($\ell_D{=}6$):
    
    \begin{equation}
        \mathbf{A_P} = \mathrm{Conv2D}\bigl(\mathrm{CrossAtt}(\mathbf{E_V}^L, \mathbf{E_T}) \bigr) \in \mathbb{R}^{B \times L \times d}
    \end{equation}
    
    Then the adaptation prompts provide to the last cross-modal decoder layer $\ell_D = 6$, as the describle in \autoref{Zero-initialized Cross-attention}.
    
\textbf{Arch.~III — Early Decoder Injection.} 
    Same as Arch.~II but insert at the \emph{first} decoder layer ($\ell_D{=}1$) for progressive refinement:
    
    \begin{equation}
    \mathbf{F}_i^{\ell_D} = \mathrm{Decoder}_{\ell_D}(\mathbf{F}_i^{\ell_D-1}, \mathbf{P}),\ \ell_D \in \{1,...,6\}
    \end{equation}

\begin{wrapfigure}[15]{R}{0.5\textwidth}
            \centering
            \includegraphics[width=1\linewidth]{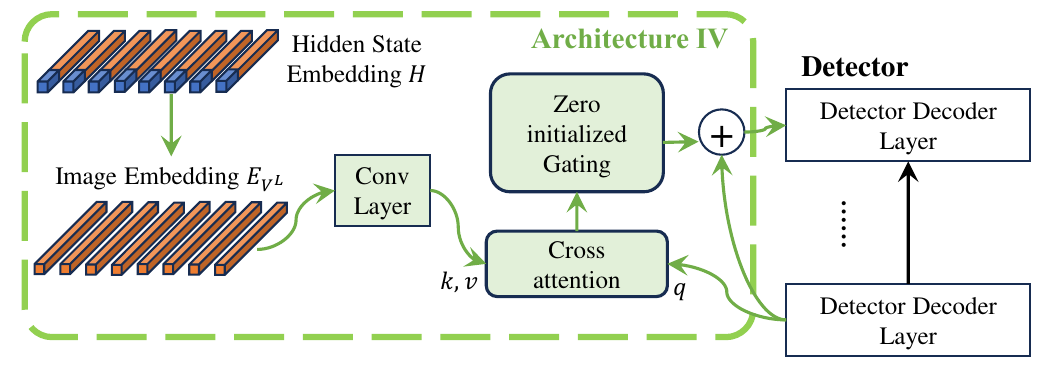}
            \caption{The adapter architecture IV of \LED: We ultimately used image embedding part as the key and value for cross-attention. The query is derived from the embedding of the last detector decoder. The out of attention to the input of the next decoder through a zero-initialized gating.}
            \label{adapter_architecture_4}
             \vspace{-8pt}
        \end{wrapfigure}

\textbf{Arch.~IV — Text-Free Adaptation.} 
    As shown in the \autoref{adapter_architecture_4}, Eliminating text embeddings $\mathbf{E_T}$, we directly project image embeddings $\mathbf{E}_v$ using the same convolutional layer as Arch. II. This tests the hypothesis that visual embeddings alone carry sufficient multimodal signals.
    
    As illustrated in the \autoref{omnilabel_evaluation_adapter_design}, Arch.~II boosts descr-AP by +4.17\% over Arch.~I but sacrifices 2.20\% categ-AP, implying that late fusion favours free-form phrases at the cost of categorical priors. 
    Early injection (Arch.~III) recovers part of the categ-AP while retaining most of the descr-AP.  
    Surprisingly, text-free Arch.~IV outperforms all others, suggesting (i) vision embeddings already encode rich cross-modal cues from pretraining, and (ii) naive text–vision fusion may introduce semantic noise without explicit contrastive alignment.

    \begin{table}[!tb]
      \renewcommand{\arraystretch}{1.1}
      \centering
      \begin{tabular}{p{0.18\columnwidth}p{0.20\columnwidth}p{0.20\columnwidth}p{0.20\columnwidth}}
        \toprule
        \textbf{Method} & \textbf{AP} descr,categ & \textbf{AP} categ & \textbf{AP} descr\\
        \midrule
        G-DINO & 21.69\% & 33.25\% & 16.09\%  \\
        \hline
        \hyperref[adapter_architecture_1]{Arch.~I} &
          22.57\% \textcolor{blue}{$\uparrow$0.88\%} &
          33.80\% \textcolor{blue}{$\uparrow$0.55\%} &
          16.95\% \textcolor{blue}{$\uparrow$0.86\%} \\
        \hyperref[adapter_architecture_23]{Arch.~II} &
          24.52\% \textcolor{blue}{$\uparrow$2.84\%} &
          31.05\% \textcolor{red}{$\downarrow$2.20\%} &
          20.26\% \textcolor{blue}{$\uparrow$4.17\%} \\
        \hyperref[adapter_architecture_23]{Arch.~III} &
          24.63\% \textcolor{blue}{$\uparrow$2.95\%} &
          32.07\% \textcolor{red}{$\downarrow$1.19\%} &
          19.99\% \textcolor{blue}{$\uparrow$3.90\%} \\
        \hyperref[adapter_architecture_4]{Arch.~IV} &
          25.98\% \textcolor{blue}{$\uparrow$4.29\%} &
          33.89\% \textcolor{blue}{$\uparrow$0.63\%} &
          21.06\% \textcolor{blue}{$\uparrow$4.97\%} \\
        \bottomrule
      \end{tabular}
      \caption{OmniLabel evaluation for different adapter architectures. Architecture I: Full cross-attention fusion; II: Conv-based prompt projection; III: Early decoder injection; IV: Text-free adaptation.}
      \label{omnilabel_evaluation_adapter_design}
    \end{table}

\subsubsection{Different LLMs}
\label{Different LLMs}

        \label{adapter_architecture_Analysis}

We employ the InternVL-1B, 2B, and 8B models to extract the hidden state from the $\ell_{LM}=$ 2nd layer of the LLM decoder to generate the adaptation prompts. InternVL-1B, 2B, and 8B models are all based on InternViT-300M as the vision encoder, and they use Qwen2-0.5B, InternLM2-1.8B, and InternLM2.5-7B as the LLMs, respectively. \autoref{omnilabel_evaluation} demonstrates the performance comparison between our method and Grounding DINO on the OmniLabel benchmark \cite{schulter2023omnilabel}.

\LED~achieves significant improvements in open-set object detection, with 6.22\% and 6.19\% AP gains for \LED~Qwen2-0.5B and \LED~+InternLM2-1.8B, respectively. The attention mechanisms in LLM's decoder layers effectively \textbf{fusing semantically knowledge} to visual features, enabling better grounding of complex object descriptions.

In particular, the description and positioning joint task shows remarkable 11.63\% and 11.53\% AP improvements, demonstrating MLLMs' capacity for spatial relationship understanding. We hypothesize that this stems from the model's strong logical reasoning capabilities, which enhance the understanding of spatial relationships in textual descriptions, such as "a person holding a phone." The visual-textual co-attention mechanisms likely help disambiguate spatial configurations that challenge pure visual encoders. This phenomenon is clearly observed in the analysis of \autoref{Case Study} (see Supplement).

\begingroup
  \setlength{\textfloatsep}{4pt}      
  \setlength{\abovecaptionskip}{8pt}
\begin{table}[!b]
\captionsetup{skip=6pt}
  \centering
  \resizebox{\columnwidth}{!}{%
    \renewcommand{\arraystretch}{1.1}
    \begin{tabular}{p{0.18\columnwidth} p{0.16\columnwidth} p{0.16\columnwidth} p{0.16\columnwidth} p{0.17\columnwidth}}
      \toprule
      \textbf{Framework} & \textbf{AP} descr,categ & \textbf{AP} categ & \textbf{AP} descr& \textbf{AP} descr-pos\\
      \midrule
      G-DINO & 21.69\% & 33.25\% & 16.09\% & 24.61\% \\
      \hline
      Qwen2-0.5B &
        27.90\%  \textcolor{blue}{$\uparrow$6.22\%} &
        33.93\% \textcolor{blue}{$\uparrow$0.67\%} &
        23.70\% \textcolor{blue}{$\uparrow$7.61\%} &
        36.24\%  \textcolor{blue}{$\uparrow$11.63\%} \\
      InternLM2-1.8B &
        27.87\% \textcolor{blue}{$\uparrow$6.19\%} &
        33.44\% \textcolor{blue}{$\uparrow$0.19\%} &
        23.90\%  \textcolor{blue}{$\uparrow$7.81\%} &
        36.13\% \textcolor{blue}{$\uparrow$11.53\%} \\
      InternLM2.5-7B &
        26.33\% \textcolor{blue}{$\uparrow$4.65\%} &
        32.03\% \textcolor{red}{$\downarrow$1.22\%} &
        22.36\%  \textcolor{blue}{$\uparrow$6.27\%} &
        34.34\% \textcolor{blue}{$\uparrow$9.73\%} \\
      \bottomrule
    \end{tabular}%
  }
  \caption{OmniLabel evaluation on GroundingDINO vs.\ adaptation prompts \LED~+ Qwen2-0.5B, \LED~+ InternLM2-1.8B and \LED~+ InternLM2.5-7B.}
  \label{omnilabel_evaluation}
\end{table}

\begin{table*}[!b]
  \centering
  \resizebox{\textwidth}{!}{%
    \renewcommand{\arraystretch}{1.1}
    \begin{tabular}{p{0.17\columnwidth}|p{0.07\columnwidth} p{0.07\columnwidth} p{0.07\columnwidth}|
                                    p{0.07\columnwidth} p{0.07\columnwidth} p{0.07\columnwidth}|
                                    p{0.07\columnwidth} p{0.07\columnwidth}}
      \toprule
       & \multicolumn{3}{c|}{RefCOCO} & \multicolumn{3}{c|}{RefCOCO+} & \multicolumn{2}{c}{RefCOCOg} \\
       & eval & test A & test B & eval & test A & test B & eval & test \\
      \cmidrule(lr){2-4}\cmidrule(lr){5-7}\cmidrule(lr){8-9}
      G-DINO &
        48.0\% & 54.8\% & 41.5\% & 48.5\% & 53.0\% & 44.1\% & 66.2\% & 66.4\% \\
      \midrule
      Qwen2-0.5B &
        51.3\% \textcolor{blue}{↑3.3\%} & 58.6\% \textcolor{blue}{↑3.8\%} & 43.4\% \textcolor{blue}{↑1.9\%} &
        50.8\% \textcolor{blue}{↑2.3\%} & 56.0\% \textcolor{blue}{↑3.0\%} & 45.3\% \textcolor{blue}{↑1.2\%} &
        70.2\% \textcolor{blue}{↑4.0\%} & 70.5\% \textcolor{blue}{↑4.1\%} \\
      InternLM2.5-7B & 
        52.4\% \textcolor{blue}{↑4.4\%} & 59.8\% \textcolor{blue}{↑5.0\%} & 45.1\% \textcolor{blue}{↑3.6\%} &
        51.8\% \textcolor{blue}{↑3.3\%} & 57.2\% \textcolor{blue}{↑4.2\%} & 46.8\% \textcolor{blue}{↑2.7\%} &
        71.2\% \textcolor{blue}{↑5.0\%} & 71.4\% \textcolor{blue}{↑5.0\%} \\
      \bottomrule
    \end{tabular}%
  }
  \caption{Performance comparison on RefCOCO, RefCOCO+, and RefCOCOg datasets with improvements over GroundingDINO.}
  \label{RefCOCO evaluation}
\end{table*}
\endgroup

Notably, category detection shows minimal improvements (0.67\% AP and 0.19\% gain) across experiments because closed-set category recognition relies primarily on visual prototype matching rather than semantic understanding. Thus, the Swin-Tiny encoder's features may already saturate performance on this conventional detection task, leaving limited room for text-guided enhancement.


We evaluated our model on RefCOCO, RefCOCO+, and RefCOCOg \cite{yu2016modeling}; the results are reported in \autoref{RefCOCO evaluation}. Together with the OmniLabel evaluations, we observe that \LED~+ Qwen2-0.5B performs on par with \LED~+ InternLM2.5-7B despite the latter having $14\times$ more parameters, indicating diminishing returns from scaling LLMs for visual grounding and echoing the hypothesis of \cite{elbayad2019depth} that small models can suffice for certain vision–language mappings. 
In \autoref{tab:d3_ovdeval_combined}, we further include D3 and OVDEval benchmarks. Integrating LED-using only the first two layers of Qwen2-0.5B-consistently improves D3 (Full/Pres/Abs) and strengthens OVDEval attributes and relations, with the largest gains in the relationship category. Overall, these results confirm that LED mainly enhances text-conditioned grounding and compositional reasoning while preserving detector efficiency.

\begin{figure}[!htb]
  \centering
  \begin{minipage}[t]{0.5\linewidth}
    \vspace{0pt}
    \includegraphics[width=\linewidth]{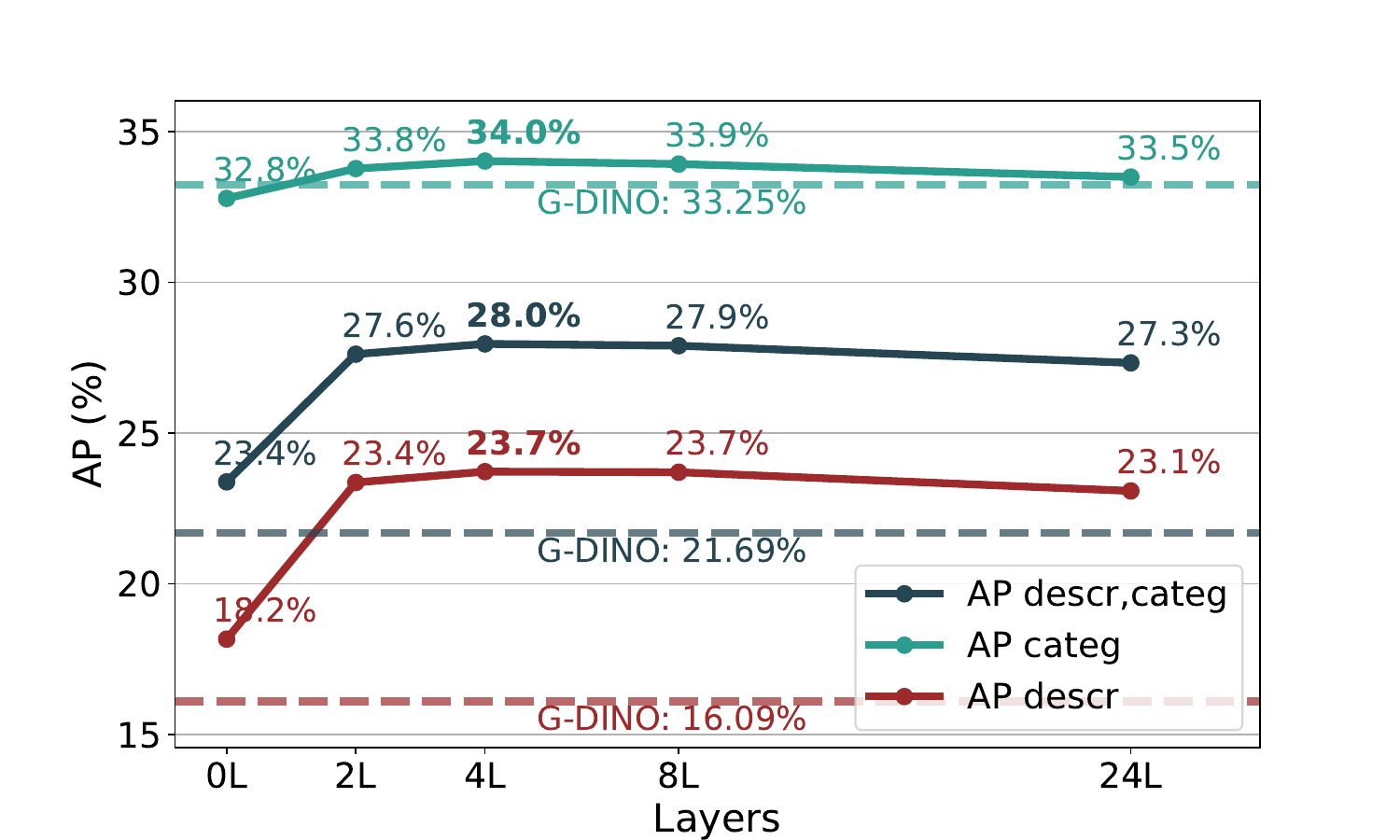}
    \captionof{figure}{OmniLabel prompts evaluated across LLM decoder layers $0$–$24$.}
    \label{Ablations: Different Decoder Layer}
  \end{minipage}\hfill
  \begin{minipage}[t]{0.48\linewidth}
    \vspace{2pt}
    \captionof{table}{OmniLabel performance (\%) of \LED~based on InternVL2-1B (IVL2-1B) compare with SOTA GLIP-T–based data-generation pipelines.}
    \label{tab:omnilabel_generation}
    \centering
    \renewcommand{\arraystretch}{1.1}
    \begin{tabular}{p{0.36\linewidth} c c c}
      \toprule
      \multirow{2}{*}{\textbf{Method}} & \textbf{AP} & \textbf{AP} \\
       & descr,categ & descr \\
      \midrule
      GLIP-T \cite{li2022grounded}         & 19.3 & 16.4 \\
      \hline
      \textsc{NEG-Text} \cite{zhao2024generating} & 22.2 & 18.8 \\
      \textsc{DesCo} \cite{li2023desco}    & 23.8 & 21.0 \\
      \textbf{\LED~IVL2-1B}                & \textbf{24.8} & \textbf{24.7} \\
      \bottomrule
    \end{tabular}
  \end{minipage}
\end{figure}

\begin{table}[t]
\centering
\small
\begin{tabular}{lccccccccc}
\toprule
\multirow{2}{*}{Method} & \multicolumn{3}{c}{D3 (AP)} & \multicolumn{5}{c}{OVDEval (AP by sub-category)} \\
\cmidrule(lr){2-4} \cmidrule(lr){5-9}
 & Full & Pres & Abs & Color & Material & Negation & Position & Relationship \\
\midrule
G-DINO & 20.7 & 20.1 & 22.5 & 8.9 & 9.9 & 23.8 & 23.8 & 27.2 \\
G-DINO + LED & \textbf{21.8} & \textbf{21.7} & \textbf{23.6} & \textbf{11.1} & \textbf{10.1} & 23.8 & 23.7 & \textbf{30.9} \\
\bottomrule
\end{tabular}
\caption{Combined results on D3 and OVDEval Sub-categories. G‑DINO-T’s pre‑training (on O365,GoldG,Cap4M) excludes OVDEval categories “celebrity” (e.g., Taylor Swift), “landmark” (e.g., the Stature of Liberty), and “logo” (e.g., McDonald's logo), so its AP is almost 0\% there and omitted for clarity.}
\label{tab:d3_ovdeval_combined}
\end{table}

\subsection{Comparison with synthetic-data baselines.}
    We integrate \LED~into the GLIP-T detector and evaluate on OmniLabel, juxtaposing it with two SOTA data-generation pipelines—NEG-Text \cite{zhao2024generating} and DesCo \cite{li2023desco}.  
    Without synthesising any extra training data, \LED~attains the best overall score and the highest descr-AP.
    
    
    As shown in the Table \ref{tab:omnilabel_generation}, \LED~achieves the highest overall score while requiring no synthetic labels, showcasing superior knowledge fusion over data-generation approaches. \textbf{Notably}, LED IVL2-1B particularly excels on the more challenging description samples, achieving a description AP of 24.7\%, which is 3.7\% higher than DesCo (21.0\%), demonstrating its superior performance in complex semantic understanding and localization.

\subsection{Deeper Decoder Layers Offer No Additional Benefit}
\label{Ablations}



To quantify the information contained at different decoder layers of the MLLM, we ablated adaptation prompts derived from each layer—including the pure ViT embeddings from the $0$ “layer” (vision encoder only, without LLM)—and evaluated them on OmniLabel (see \autoref{Ablations: Different Decoder Layer}). The ViT-only prompts yield minimal gains in the categorical AP, whereas prompts from early LLM decoder layers remain rich in semantic cues. In InternVL2-1B (24-layer Qwen2-0.5B), using only the first two decoder layers already yields significant improvements, with performance peaking at layer 4. \textbf{Notably}, deeper decoder layers do not guarantee better features. This aligns with our attention-map analysis in \autoref{Methodology} and \autoref{attention_map}(see Supplementary), which shows that image processing completes in early layers, “condensing” most visual information into those representations. 

\subsection{Detector with \LED~\textit{vs.}\ Direct-LLM Substitution}
As shown in the \autoref{hidden-state substitution}(see Supplementary), replacing the detector’s visual embeddings with raw hidden states extracted from a frozen MLLM not only fails to match the original detector’s performance but can even cause a dramatic collapse in detection accuracy, confirming that latent features alone are neither geometrically nor semantically aligned with the detector’s visual space. Effective transfer therefore requires an explicit knowledge-fusion mechanism that jointly conditions on both language and vision, and the specialized detector \textbf{remains indispensable} for reliable performance in vision–language tasks.

\subsection{Detector with \LED~\textit{vs.}\ Large VLMs}

We benchmark the InternVL2 family on OmniLabel and D3 and contrast them with lightweight detectors. While InternVL2-8B and InternVL2-76B exhibit strong semantic capacity, they require tens to hundreds of times more parameters and GFLOPs than GLIP-T and G-DINO. By comparison, LED, built on the first two layers of Qwen2-0.5B, adds only 8.7\% FLOPs to the detector while providing effective semantic fusion. As \autoref{tab:sota_vlm_led}, LED markedly improves G-DINO/GLIP-T at a tiny computational cost, whereas scaling to very large MLLMs is far less efficient for the detection setting. In short, MLLMs contribute generality and rich semantics but are not optimized for efficiency or detection; classic detectors are efficient but semantically shallow. LED bridges the gap by injecting LLM semantics without sacrificing detector efficiency.

\begin{table}[t]
\centering
\small
\begin{tabular}{lcccc}
\toprule
Model & Params & GFLOPs & AP categ & AP descr \\
\midrule
OmDet & 610M & 640G & 13.3\% & 19.5\% \\
Qwen2.5-VL-3B & 3.0B & 5.44T & 4.2\% & 4.8\% \\
InternVL2-8B & 8.1B & 8.47T & 4.5\% & 5.2\% \\
InternVL2-76B & 75B & 17.92T & 15.3\% & 19.3\% \\
\hline
GLIP-T & 232M & 434G & 16.4\% & 25.8\% \\
GLIP + LED & 290M & 472G & \textbf{24.7\%} & \textbf{36.8\%} \\
G-DINO & 172M & 412G & 16.1\% & 24.6\% \\
G-DINO + LED & 230M & 448G & \textbf{23.7\%} & \textbf{36.2\%} \\
\bottomrule
\end{tabular}
\caption{Comparison with SOTA detectors and large VLMs using LED. LED adds a small overhead (first two layers of Qwen2-0.5B; $\sim$8.7\% FLOPs) yet yields strong gains, especially on description-style queries.}
\label{tab:sota_vlm_led}
\end{table}

%% file: sec/5_conclusion.tex
\section{Conclusions}
\label{sec:concl}

Our work introduces a systematic framework that leveraging early MLLM decoder layers to boost visual grounding efficiency. We (1) demonstrate that shallow Transformer hidden states preserve rich spatial–semantic correlations; (2) design a zero-initialized cross-attention adapter for seamless MLLM-to-detector knowledge transfer; and (3) develop two adaptation-prompt schemes that sharpen descriptive precision while maintaining category accuracy. Integrating Qwen2-0.5B with Swin-Tiny yields a 3.82\% OmniLabel gain at just 8.7\% extra GFLOPs, while swapping in InternVL ViT pushes the improvement to 6.22\%, alongside notable RefCOCO/+/g boosts. Extensive ablation studies across adapter variants, LLM scales, and decoder layers \textbf{confirm} our method’s robustness—especially for detailed and zero-shot queries. Our generalizable adapter design supports real-time vision–language applications and mitigates overfitting due to manual prompt bias, thereby advancing practical MLLM deployment.

%% file: sec/X_suppl.tex
\clearpage
\setcounter{page}{1}

\section{Experiment Details}

\subsection{Attention Maps in Qwen2-0.5B Decoding}
\label{Attention Maps in Qwen2-0.5B Decoding}
    \begin{figure*}[!ht]
        \centering
        \includegraphics[width=1\linewidth]{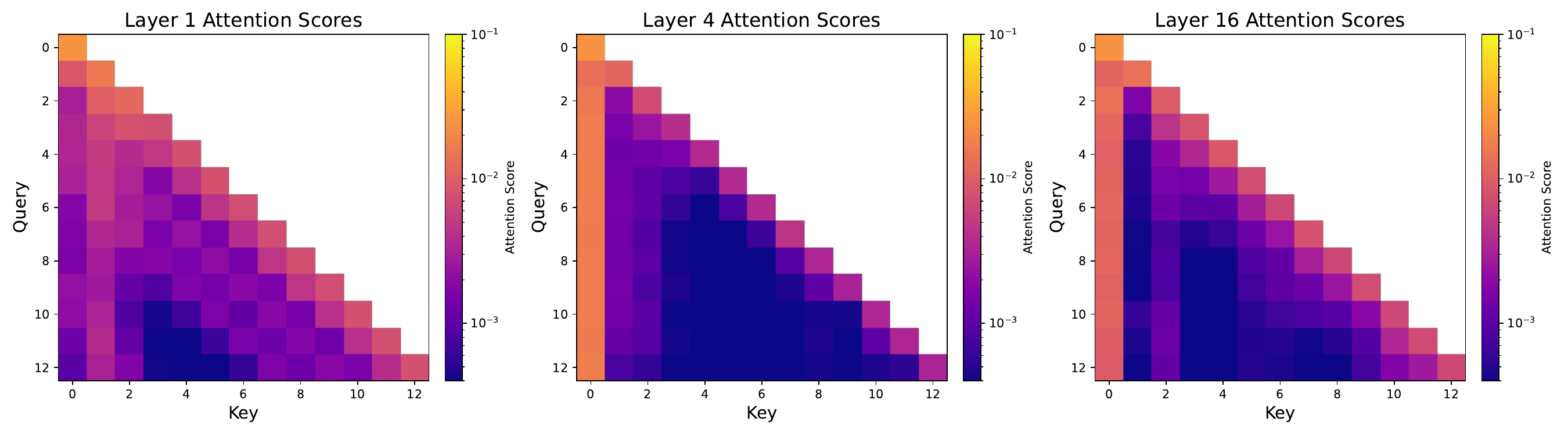}
        \caption{Attention maps during the decoding process of Qwen2-0.5B.}
    \label{attention_map}
    \vspace{-8pt}
    \end{figure*}

    Based on the attention maps during the decoding process of Qwen2-0.5B, We draw the attention scores for different layer as fellow:
    \begin{equation}
    \text{Attention Score}(Q, K) = \frac{QK^T}{\sqrt{d_k}}
    \end{equation}
    
    Where, $Q, K$ are the Query and Key matrix in the transformer Decoder layer, $\text{softmax}(\cdot)$ .as show in \autoref{attention_map}, we have reached an approximate conclusion \cite{chen2024imageworth12tokens}. We can observe that in the first layer, attention is distributed relatively smoothly across different types of tokens. In the deeper layers, starting from local attention, attention scores are aggregated onto system prompts, instructions, and output tokens, while attention to image tokens becomes quite sparse. In the deeper layers, there are strong vertical lines (in the system prompts) that dominate most of the attention scores. The presence of these strong vertical lines indicates that certain input tokens remain highly attended to throughout the decoding process. If large-scale, high-intensity attention to image tokens were also maintained in the deeper layers, it would suggest that a significant amount of visual information is still needed for inference at those stages. However, as seen in the visualization, the deeper layers primarily focus on some key text tokens, which precisely indicates that the processing of images has already been completed in the earlier layers, and most of the image information has been "condensed" into the representations.

\subsection{Adapter Architecture Detail}
\label{Adapter Architecture Detail}

    \begin{wrapfigure}{R}{0.5\textwidth}
        \centering
        \includegraphics[width=1\linewidth]{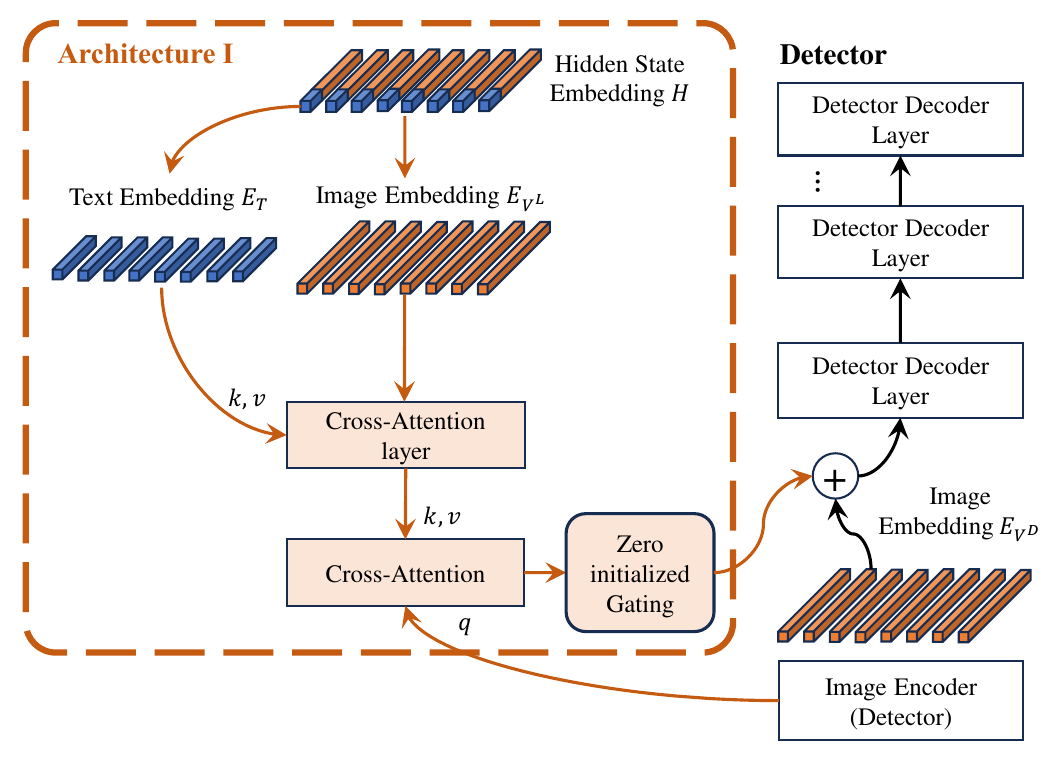}
        \caption{The adapter architecture $\mathrm{I}$ of our approach.}
        \label{adapter_architecture_1}
        \vspace{-8pt}
    \end{wrapfigure}

    \textbf{Architecture I (Double Cross-Modal Fusion).}
    As show in \autoref{adapter_architecture_1}, this design leverages two sequential cross-attention layers to incorporate both textual and visual cues into the Detector. First, the image embeddings $\mathbf{E_V}_L$ attend to text embeddings $\mathbf{E_T}$, ensuring high-level semantic alignment via a \emph{text-guided image fusion}. Second, the fused features further guide the Detector through a \emph{prompt-enhanced detection} stage, where Detector queries attend to the fused representation. A zero-initialized gating mechanism adaptively combines these cross-modal features back into the Detector’s backbone.

    \begin{wrapfigure}{R}{0.5\textwidth}
        \centering
        \includegraphics[width=1\linewidth]{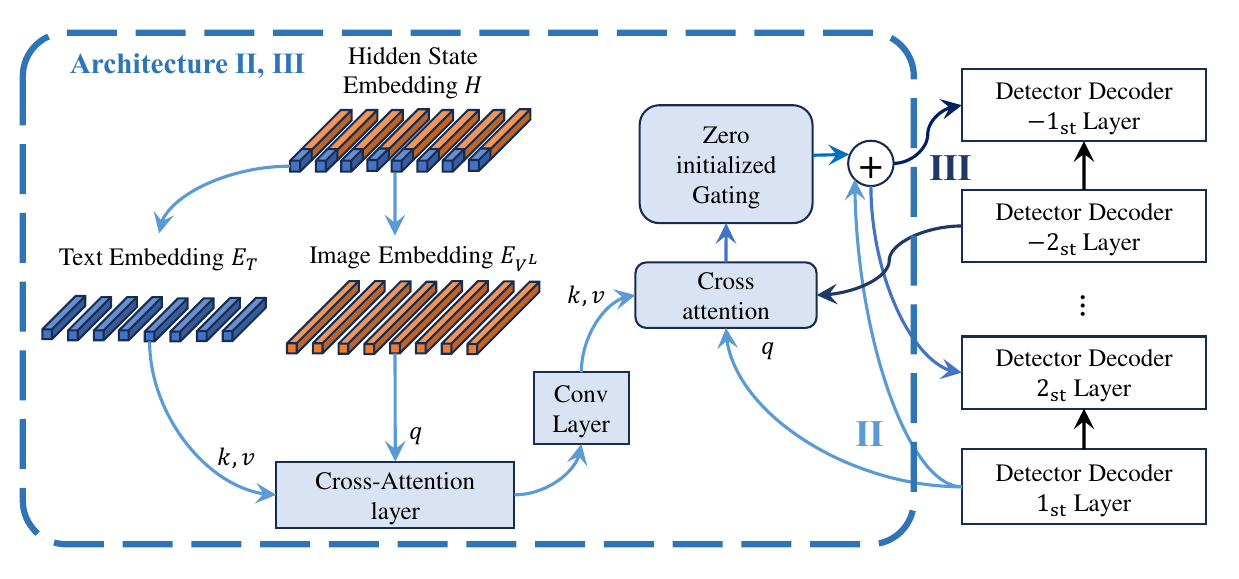}
        \caption{The adapter architecture $\mathrm{II}$, $\mathrm{III}$ of our approach.}
        \label{adapter_architecture_23}
        \vspace{-8pt}
    \end{wrapfigure}

    \textbf{Architecture $\mathrm{II}$, $\mathrm{III}$ (Late Prompt Projection).} As show in \autoref{adapter_architecture_23}, we remove the second cross-attention step to streamline the flow into the Detector. A $3\times 3$ convolution directly transforms the fused adaptation prompts into the compatible feature dimension before feeding them into the final cross-modal decoder.

\subsection{Experiment Configuration}
\label{sec:Experiment Configuration}
    All experiments are conducted on eight NVIDIA A100 GPUs.  
    During Stage 1 (pre-training) and Stage 2 (fine-tuning), we train the MLP layer for exactly one epoch each.  
    For pre-training we use a batch size of 128, a learning rate of $1\times10^{-3}$, weight decay of 0.03, and no warm-up.  
    For fine-tuning, the warm-up ratio is set to 0.03, while the batch size, learning rate, and weight decay are fixed at 128, $4\times10^{-5}$, and 0, respectively.  
    In Stage 3 we train the adapter for one epoch with a batch size of 64, employing cosine learning-rate scheduling (base learning rate $2\times10^{-4}$) without warm-up or weight decay.  
    The MLP layer and the LLM are co-trained with the adapter using a separate learning rate of $4\times10^{-5}$.
    
    The hyper-parameters for Stages 1–3 are summarised in \autoref{Vision Decoder Share from Detector Config}.  
    To investigate the impact of model scale, we vary LLM parameter sizes as detailed in \autoref{Different MLLM Config}, with the results reported in \autoref{Different LLMs}.  
    \autoref{Different adapter Config} lists the settings used to evaluate alternative adapter architectures; the corresponding OmniLabel scores appear in \autoref{omnilabel_evaluation_adapter_design}.  
    Finally, \autoref{Ablations: Different Decoder Layer Config} specifies the setup for examining different decoder layers, and the ablation results are presented in \autoref{Ablations}.

    \begin{table}[!ht]
      \centering
      \begin{tabular}{cc}
        \begin{subtable}[t]{0.48\textwidth}
          \centering
          \small
          \begin{tabular}[t]{p{0.26\textwidth}|p{0.66\textwidth}}
            \hline
            \multicolumn{2}{c}{\textbf{(a) Vision Decoder Share from Detector}} \\
            \hline
            MLLM                         & InternVL2-1B                     \\
            $\ell_{LM}$ & 2                             \\ 
            $\mathbf{AP}$ to $\ell_D$     & 6                             \\ 
            Train Dataset             & Object365, COCO, Flickr30k      \\
            Stage 1 $lr$                 & MLP: $1e^{-3}$                  \\
            Stage 2 $lr$                 & MLP: $2e^{-5}$                  \\
            Stage 3 $lr$                 & MLP \& Adapter:$2e^{-4}$ \\
            Adapter arch.     & IV                              \\
            \hline
          \end{tabular}
          \caption{Experiment configuration details for vision decoder share from detector.}
          \label{Vision Decoder Share from Detector Config}
        \end{subtable}
        &
        \begin{subtable}[t]{0.48\textwidth}
          \centering
          \small
          \begin{tabular}[t]{p{0.26\textwidth}|p{0.58\textwidth}}
            \hline
            \multicolumn{2}{c}{\textbf{(b) Different MLLMs}} \\
            \hline \\
            MLLM                         & InternVL2: 1B, 2B, 8B            \\
            $\ell_{LM}$ & 2                             \\ 
            $\mathbf{AP}$ to $\ell_D$     & 6                             \\ 
            Train Dataset             & Object365, COCO, Flickr30k      \\
            Adapter arch.     & IV   \\
            \\ \\ 
            \hline
          \end{tabular}
          \caption{Experiment configuration details for different LLMs.}
          \label{Different MLLM Config}
        \end{subtable}
        \\[1em]
        \begin{subtable}[t]{0.48\textwidth}
          \centering
          \small
          \begin{tabular}[t]{p{0.26\textwidth}|p{0.66\textwidth}}
            \hline
            \multicolumn{2}{c}{\textbf{(c) Adapter Architectures}} \\
            \hline
            MLLM                         & InternVL2-2B                    \\
            $\ell_{LM}$ & 8                             \\ 
            $\mathbf{AP}$ to $\ell_D$     & 1, 6                          \\ 
            Train Dataset             & COCO, Flickr30k                 \\
            Adapter arch.     &  I,II,III,IV   \\
            \hline
          \end{tabular}
          \caption{Experiment configuration details for different adapter architectures.}
          \label{Different adapter Config}
        \end{subtable}
        &
        \begin{subtable}[t]{0.48\textwidth}
          \centering
          \small
          \begin{tabular}[t]{p{0.26\textwidth}|p{0.58\textwidth}}
            \hline
            \multicolumn{2}{c}{\textbf{(d) Ablations: Decoder Layers}} \\
            \hline
            MLLM                         & InternVL2-1B                    \\
            $\ell_{LM}$ & -1, 2, 4, 8, 24              \\ 
            $\mathbf{AP}$ to $\ell_D$     & 6                             \\ 
            Train Dataset             & Object365, COCO, Flickr30k      \\
            Adapter arch.     & IV    \\
            \hline
          \end{tabular}
          \caption{Ablation details for different decoder layers.}
          \label{Ablations: Different Decoder Layer Config}
        \end{subtable}
      \end{tabular}
      \caption{Experiment configuration summary.}
      \label{Experiment configuration summary}
    \end{table}

\subsection{Dynamic Image Processing}

    \begin{wrapfigure}[11]{R}{0.5\textwidth}
        \centering
        \includegraphics[width=1\linewidth]{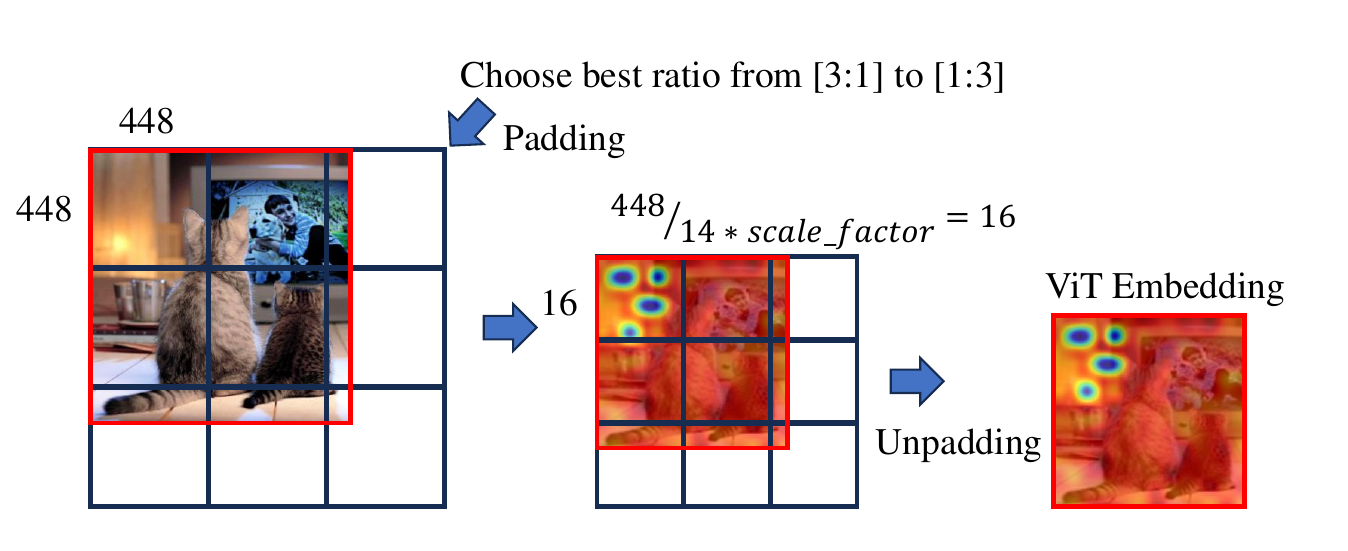}
    \caption{Dynamic Image Processing}
    \label{any_res}
    \vspace{-8pt}
    \end{wrapfigure}
    
    In \autoref{Different LLMs} and \autoref{Ablations}, Since the MLLM and Detector's Image encoder process the same image in a single inference, and considering that the Hidden state cropped from the Vision token positions in LLMs corresponds to the Vision Embedding processed by the detector, we are inspired by LLaVA\cite{liu2023visual}'s approach to dynamic image processing. The Vision Embedding sent to LLMs is processed as shown in \autoref{any_res}. For a given image, based on a $448 \times 448$ patch, rectangles are generated from a 1:3 to 3:1 aspect ratio and matched to the image. The image is then padded to the smallest rectangle that can contain the original image and sent to the Vision Transformer (ViT) of LLMs. The size of the patch embedding should be $448/(14 \times \text{scale factor}=2)=16$. Finally, the ViT embedding is cropped according to the previous padding ratio to restore the original image's aspect ratio.

\subsection{MLLM Alignment (Swin-T \& Qwen2-0.5B)}
\label{MLLM Alignment (Swin-T & Qwen2-0.5B)}

    \begin{table}
        \centering
        \begin{tabular}{lcc}
        \hline
        \textbf{MME} & \textbf{InternVL2-1B} & \textbf{Swin-T \& Qwen2-0.5B} \\
        \hline
        \textbf{total} & 1363 & 1027 \\
        \hline
         existence & 180.0 & 158.3 \\
         count & 118.3 & 105.0 \\
        position & 126.7 & 66.7 \\
         color & 135.0 & 128.3 \\
        posters & 110.2 & 75.9 \\
        celebrity & 146.8 & 131.8 \\
         scene & 148.5 & 120.8 \\
         landmark & 132.5 & 83.3 \\
         artwork & 140.0 & 80.0 \\
        OCR & 125.0 & 77.5 \\
        \hline
         \textbf{total} & 419.29 & 273.57 \\
        \hline
         commonsense reasoning & 99.29 & 68.57 \\
         numerical calculation & 62.50 & 62.50 \\
         text translation & 162.50 & 70.00 \\
         code reasoning & 95.00 & 72.50 \\
        \hline
        \end{tabular}
        \caption{Performance comparison of InternVL2-1B, Swin-T \& Qwen2-0.5B on MME Perception and Cognition tasks.}
        \label{MME eval}
    \end{table}
    
    We evaluated the alignment of Swin-T \& Qwen2-0.5B on MME and compared it with the original method of InternVL2-1B (InternVL Vit \& Qwen2-0.5B) as show in \autoref{MME eval}. To evaluate the impact of replacing the vision encoder and aligning the LLM on MLLM performance, we compared the score gaps between InternVL2-1B and Swin-T \& Qwen2-0.5B on the MME benchmark. Compared to InternVL2-1B, Swin-T \& Qwen2-0.5B scored 1027 on perception, a decrease of 335 (InternVL2-1B scored 1362.97), and 273.57 on cognition, a decrease of 145.72 (InternVL2-1B scored 419.29). It is worth noting that, although more advanced alignment strategies may exist, our work does not aim to achieve maximum alignment between an extremely lightweight vision encoder (Swin-Tiny) and the LLM, particularly without employing strategies such as Dynamic Tiling.

\section{Why MLLM Hidden States Cannot Replace Visual Detectors}
\label{hidden-state substitution}
    One might be tempted to ask: \emph{“Can we simply replace a specialized visual detector with a LLM, MLLM or VLM and still achieve comparable performance?”} To answer this question, Table~\ref{tab:detector_vs_llm} presents a head-to-head comparison between our detector-based pipeline and an end-to-end LLM approach across multiple benchmarks. As the results make clear, relying solely on the LLM without a dedicated object detector leads to a substantial drop in accuracy and consistency.
    \begin{table}[ht]
      \centering
    \caption{\textbf{Detector vs.\ MLLM Hidden-State Features on COCO-2017 \textit{val}.} Baselines are two GroundingDINO variants and vision-only SwinTiny-BERT; substitutions plug frozen decoder states from LLaVA-1.5 (L8) or InternVL2-2B (L2/L8).}
      \label{tab:ablation}
      \begin{tabular}{@{}lccc@{}}
        \toprule
        Method & AP$_{50:95}$ & AP$_{50}$ & AP$_{75}$\\
        \midrule
        \textsc{GroundingDINO-Pre}          & 31.23 & 44.37 & 34.09\\
        \textsc{GroundingDINO-COCO SFT}         & 57.23 & 73.27 & 63.18\\
        \textsc{SwinTiny-BERT}      & 41.97 & 57.45 & 45.85\\
        \textsc{LLaVA-L8}           & 24.19 & 39.47 & 24.94\\
        \textsc{InternVL2-L2}            & 38.28 & 55.25 & 41.18\\
        \textsc{InternVL2-L8}            & 42.64 & 63.77 & 44.92\\
        \bottomrule
      \end{tabular}
      \label{tab:detector_vs_llm}
    \end{table}

    \textsc{GroundingDINO-Pre} establishes a detector base on  Open-GroundingDINO \cite{liu2024grounding} that has \emph{never} been exposed to COCO, being trained only on Objects365, GoldG, and Cap4M; \textsc{GroundingDINO-COCO~SFT} fine-tunes the same weights on COCO
    \textsc{SwinTiny-BERT} serves as a \emph{vision-only} control whose Swin-Tiny backbone and BERT text branch use only same framework as Open-GroundingDINO \cite{liu2024grounding} is identical to that used in the multi-modal runs yet, like the MLLMs, has never seen COCO or any object-detection data.  
    The three substitution variants—\textsc{LLaVA-L8}, \textsc{IVL2-L2}, and \textsc{IVL2-L8}—replace the Swin feature map with hidden states taken from decoder layers~2 or~8 of LLaVA-1.5 or InternVL-2B, enabling a direct test of whether raw MLLM representations can stand in for detector-oriented visual embeddings.

    As show int the \autoref{tab:detector_vs_llm}, fine-tuning the vision backbone on COCO (\textsc{GDINO-COCO}) improves AP$_{50:95}$ by \textbf{+26} points over its out-of-domain counterpart, underscoring the importance of spatial alignment learned from explicit detection supervision.  
    By contrast, treating language–vision model (MLLM) activations as a drop-in surrogate for visual tokens is ineffective: the strongest variant (\textsc{IVL2-L8}) still trails the fully visual \textsc{SwinTiny-BERT} by \textbf{$-0.67$} AP$_{50:95}$ and remains \textbf{$-14.6$} points below the COCO-tuned baseline. The gap widens to \textbf{$-33$} points when substituting \textsc{LLaVA-L8}, revealing that decoder-layer semantics alone carry minimal localisation cues.
    
    This discrepancy arises because MLLM hidden states are produced after heavy token mixing and rotary positional encoding optimised for caption generation; they lack the multi-scale geometry and inductive biases embedded in convolutional or ViT backbones.  Without an explicit fusion mechanism that jointly conditions on language and vision, the detector cannot recover accurate object coordinates, leading to systematic localisation failure.  The comparison between \textsc{SwinTiny-BERT} (whose Swin weights, like the MLLMs, have \emph{never} seen COCO) and the substitution runs highlights the central claim of this study: \textbf{knowledge fusion is indispensable—naive replacement of visual embeddings with isolated MLLM features is not a viable route to open-vocabulary detection.}

\section{Case Study}
    \label{Case Study}
\begin{figure}[!htbp]
        \centering

        \includegraphics[width=1\linewidth]{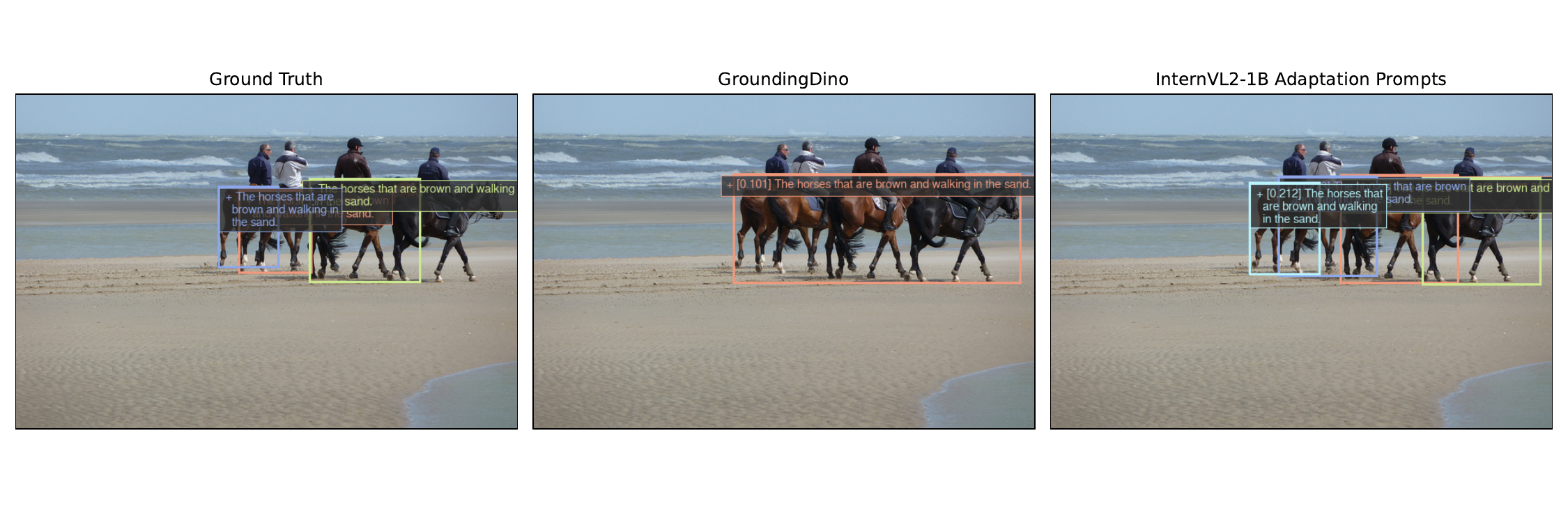}\\
        \includegraphics[width=1\linewidth]{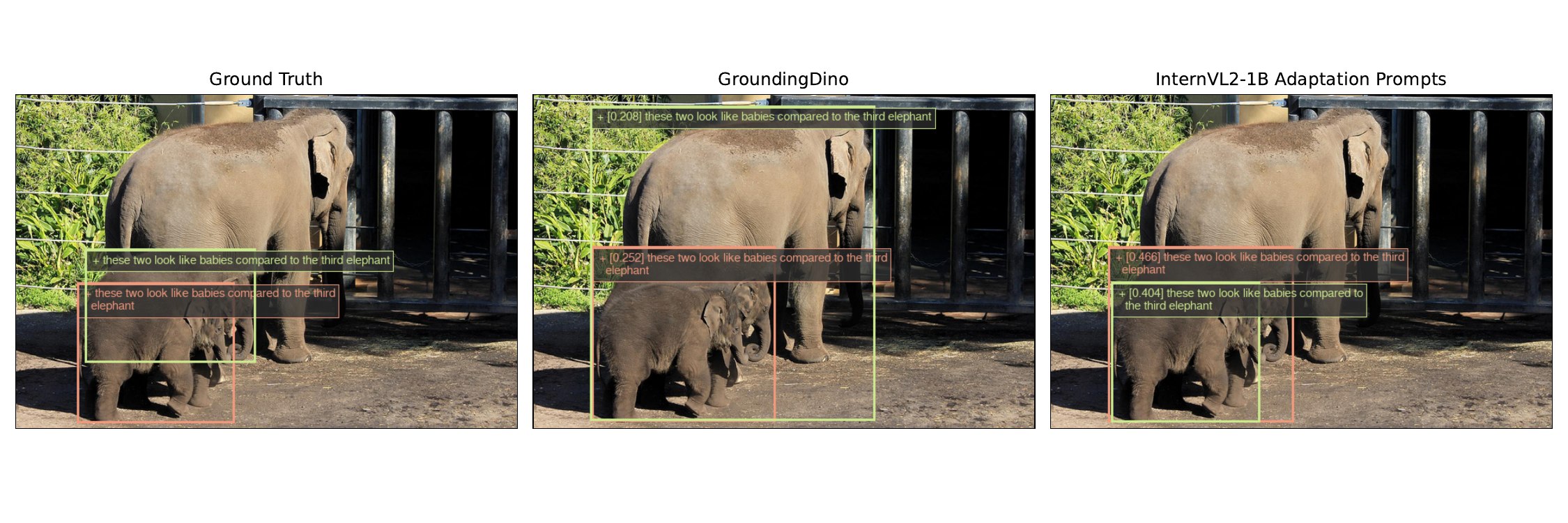}\\
        \includegraphics[width=1\linewidth]{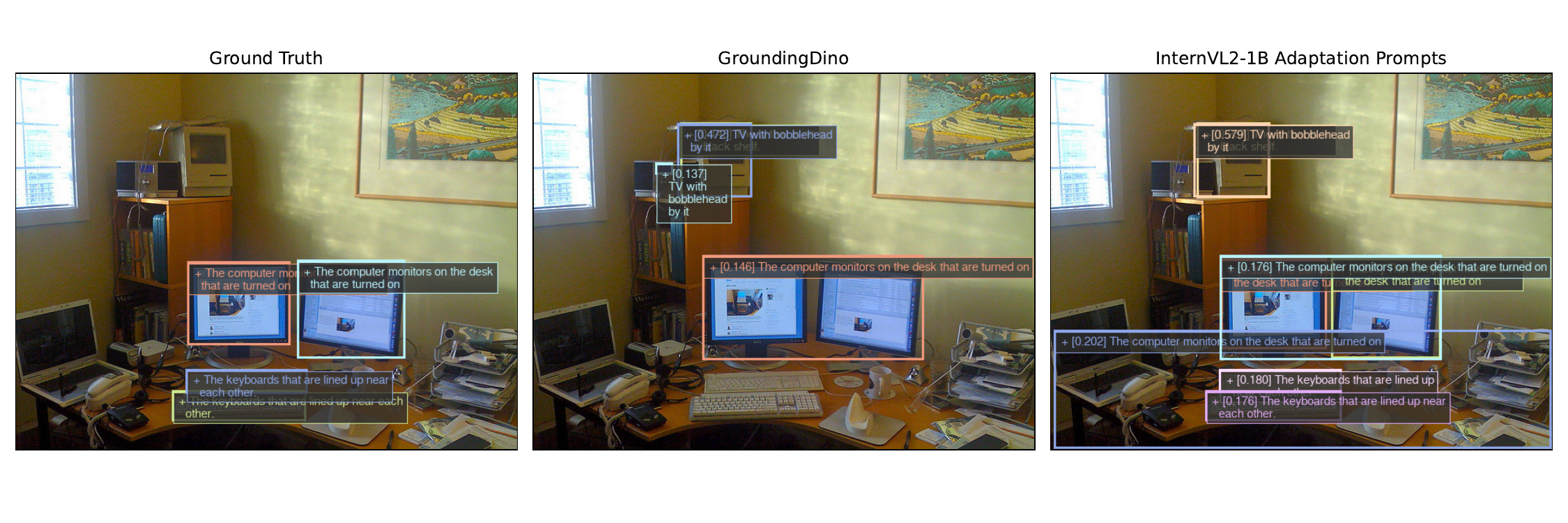}\\
        \includegraphics[width=1\linewidth]{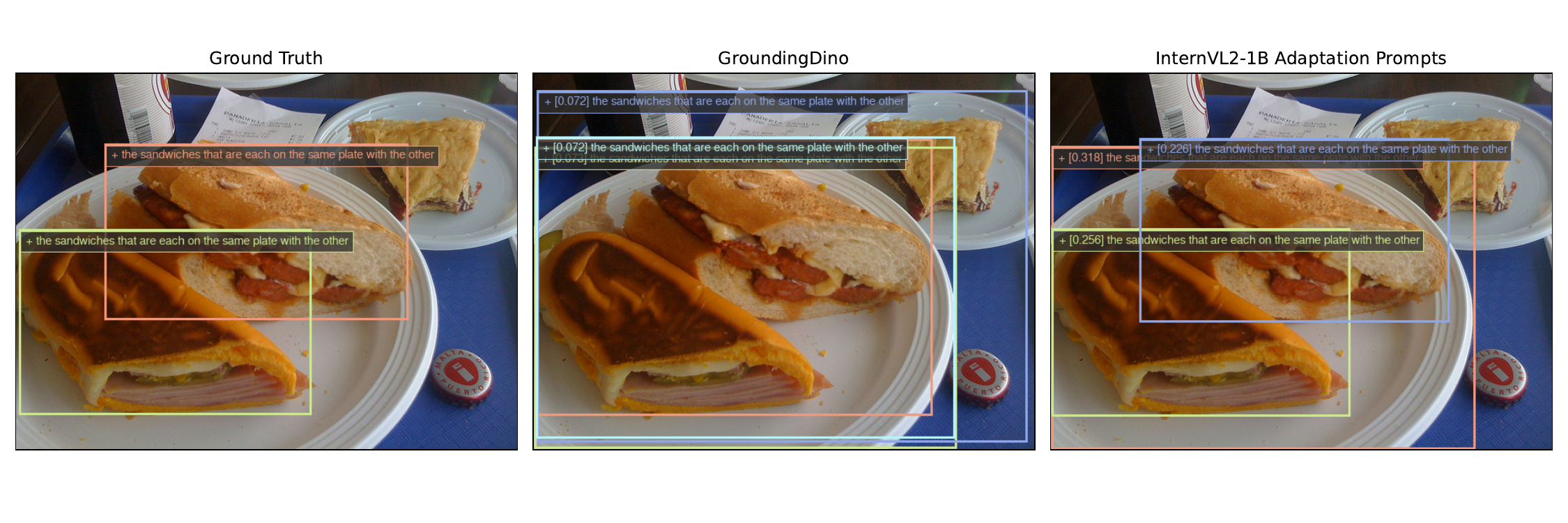}\\
        \includegraphics[width=1\linewidth]{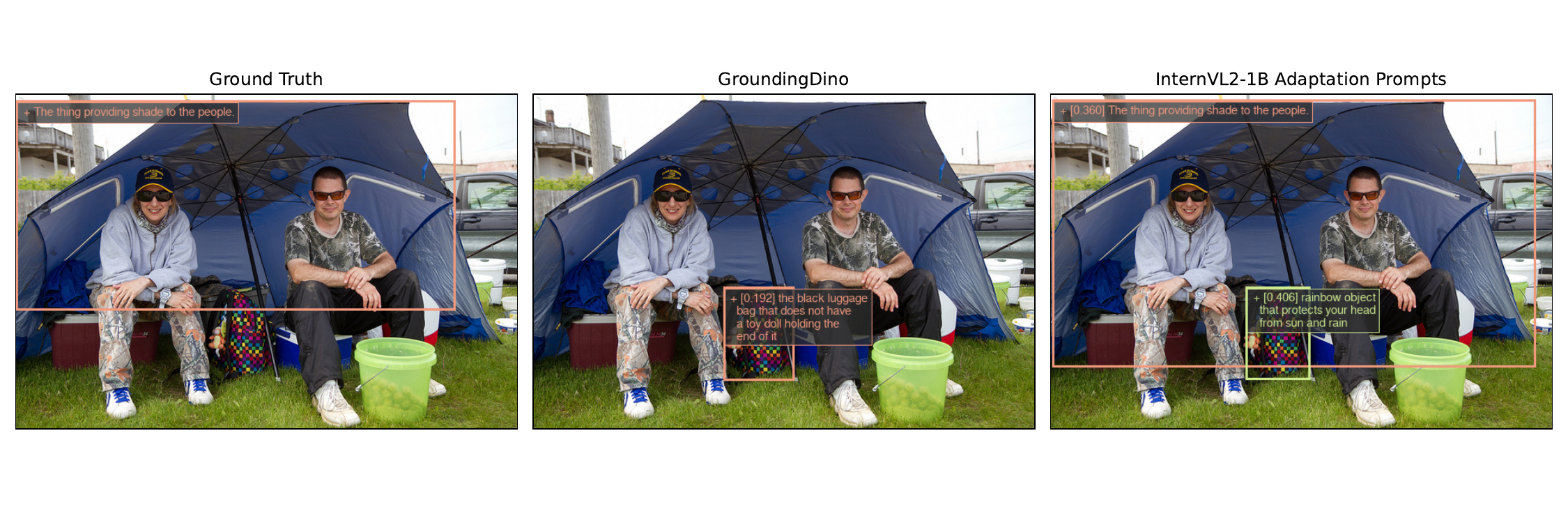}\\
        \includegraphics[width=1\linewidth]{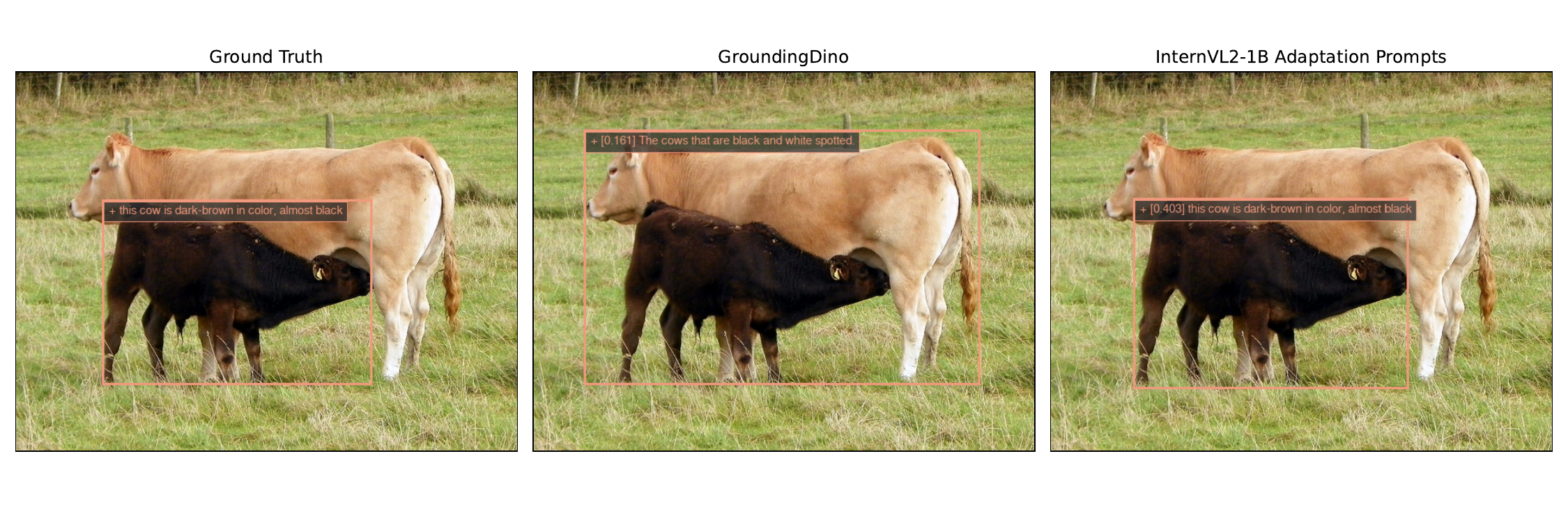}\\
        \caption{Case Study for OmniLabel using InternVL2-1B to provide adaptation prompts versus directly using GroundingDINO.}
        \vspace{-8pt}
        \label{study_case}
    \end{figure}

    We compared test results on OmniLabel using InternVL2-1B for adaptation prompts against directly applying GroundingDINO on selected samples. We excluded all category-based targets and focused solely on the detection performance of descriptive targets. As shown in \autoref{study_case}, GroundingDINO struggles to understand target descriptions involving quantities, states, ages, colors, or spatial relationships. However, when the LLM provides semantic-level understanding, the detector's ability to recognize such targets improves significantly.
    
    Notably, the adaptation prompt functions as expected, guiding the detector rather than allowing the adapter to dominate the grounding task. As seen in the recognition results in \autoref{study_case}, GroundingDINO often fails to locate individuals, instead detecting entire objects, as in the case of "The computer monitors on the desk that are turned on." Additionally, it struggles to interpret positional relationships in text descriptions, such as "the sandwiches that are each on the same plate with the other." With the adaptation prompt, the detector retains an overall understanding while incorporating individual positional relationships. Furthermore, it corrects errors in spatial logic relationships.

\section{Implementation Details}
\label{sec:Implementation Details}

\subsection{Dataset Introduction}
    \label{sec:Dataset Introduction}
    
    \autoref{Dataset Introduction} provides the attributes, sizes, and functional descriptions of all datasets used in our stages 1, 2, 3 and evaluation, to illustrate the support they offer during the process. All datasets are distributed under licences that permit non-commercial academic research.  
    
    \begin{table*}[!ht]
    \centering
    \caption{Introduction of datasets used in stage 1, stage 2, and stage 3.
    \label{Dataset Introduction}}
    \begin{tabularx}{\textwidth}{p{2.5cm}X}
    \toprule
    \textbf{Dataset} & \textbf{Introduction} \\
    \midrule
    
    \multicolumn{2}{l}{\textit{Pretraining Data for Stage 1}} \\
    \midrule
    LAION-CC-SBU 558K & The LAION-CC-SBU dataset is a curated subset of the LAION, Conceptual Captions (CC), and SBU datasets. It comprises 558K image-caption pairs for the pretraining stage for feature alignment in visual instruction tuning. \cite{liu2023visual}\\
    
    \midrule
    \multicolumn{2}{l}{\textit{Instruction Tuning Data for Stage 2}} \\
    \midrule
    ShareGPT4V & ShareGPT4V dataset is curated from LAION, CC, SBU, SAM, COCO, web-landmark, web-celebrity, wikiart, etc, resulting in total 102K high-quality image-text pairs with the help of powerful GPT4-Vision \cite{chen2023sharegpt4v}. \\
    SFT & SFT dataset comprises approximately 665K multimodal instruction-following samples, facilitating improved alignment of visual-language models to human instructions \cite{chen2023sharegpt4v}. \\
    ChartQA & ChartQA is a domain-specific visual question-answering dataset containing 18K samples designed explicitly for interpreting various types of charts, including bar graphs, pie charts, and line plots. \cite{masry-etal-2022-chartqa} \\
    AI2D & AI2D includes approximately 12K annotated diagrams paired with structured question-answer data, particularly aimed at evaluating multimodal reasoning over scientific diagrams. \cite{kembhavi2016diagram}\\
    DocVQA & DocVQA contains 10K samples that involve complex question-answering tasks over visually rich document images, emphasizing text recognition, layout analysis, and semantic comprehension. \cite{mathew2021docvqa} \\
    
    \midrule
    \multicolumn{2}{l}{\textit{Grounding Data for Stage 3}} \\
    \midrule
    Objects365 & Objects365 is a large-scale object detection dataset featuring over 1.7 million images with dense annotations covering 365 common object categories, enhancing general object detection capabilities. \cite{shao2019objects365} \\
    COCO2017 & COCO2017 provides around 118K training images annotated for object detection, and captioning tasks, widely used as a benchmark in computer vision research. \\
    Flickr30k & Flickr30k is a standard multimodal dataset with 31K images, each annotated with five descriptive captions, commonly utilized for improving image captioning and cross-modal retrieval tasks. \cite{young2014image}\\
    
    \midrule
    \multicolumn{2}{l}{\textit{Evaluation Data}} \\
    \midrule
    RefCOCO/+/g & The RefCOCO/+/g datasets consist of referring expression comprehension tasks, where models must identify objects in images based on natural language descriptions. REFCOCO has approximately 142K referring expressions, while RefFCOCO+ and RefCOCOg provide more challenging and generalized scenarios. \cite{yu2016modeling}\\
    Omnilabel & Omnilabel is a multimodal benchmark dataset containing diverse visual-language tasks designed to comprehensively evaluate the generalization and zero-shot capabilities of visual-language models across various tasks and domains. \cite{schulter2023omnilabel}\\
    
    \bottomrule
    \end{tabularx}
    \end{table*}

\subsection{Model Introduction}
    \label{Model Introduction}
    To foster clarity and reproducibility, we first outline the {\em functional role} of every third-party model or code base incorporated into LED (Table~\ref{tab:model_intros}).
    
    \begin{table}[!ht]
    \centering
    \caption{Concise introductions for third-party models and code bases.}
    \label{tab:model_intros}
    \begin{tabularx}{\textwidth}{l X}
    \toprule
    \textbf{Model / Code} & \textbf{Introduction} \\
    \midrule
    Qwen2-0.5B & A 0.5-billion-parameter decoder-only LLM trained on $\sim$2 T multilingual tokens; we tap its early hidden states for knowledge fusion in Stage~3 of LED \cite{qwen2}. \\
    
    InternVL2 (1 B / 2 B / 8 B) & A family of vision–language foundation models pretrained on 20 M image–text pairs; the 1 B and 2 B variants are used as alternative language decoders in our ablations~\cite{zhao2022exploiting,minderer2023scaling,li2023desco,yao2024detclipv3,zhao2024taming}. \\
    
    LLaVA-1.5 & An open-source multimodal chat model aligning a CLIP vision encoder with a Vicuna language head via instruction tuning; included as an external baseline for grounding performance \cite{liu2024visual}. \\
    
    Open-GroundingDINO & An open-vocabulary detector that couples a Swin-Tiny backbone with a text encoder; serves as the base detector into which our LED adapters are inserted \cite{liu2024grounding}. \\
    
    Swin-Tiny & A hierarchical vision encoder pretrained on ImageNet-22K; provides four-stage feature maps consumed by the detector backbone \cite{liu2021Swin}. \\
    \bottomrule
    \end{tabularx}
    \end{table}



\section{Broader Impacts}
    \label{sec:broader_impacts}
    \vspace{-0.5em}
    
    \paragraph{Potential benefits.}
    By enabling {\em lightweight open-vocabulary grounding} with minor extra FLOPs, LED can  
    (i) improve on-device perception for assistive robotics and smart prosthetics;  
    (ii) lower the computational barrier for researchers in low-resource regions to experiment with vision–language models; and  
    (iii) accelerate scientific discovery in ecology, astronomy, and digital humanities, where long-tail object categories are common.
    
    \paragraph{Potential harms.}
    Hallucinating an object that is not present—could propagate misinformation through downstream captioning or retrieval systems.  
    Biases inherited from pre-training corpora may yield disparate false-positive rates across demographic groups, disproportionately affecting already-marginalised communities.
    
    \paragraph{Mitigation strategies.}
    \textbf{More advanced models and algorithms.} As the LLM field continues to advance, the Hallucinating problem will continue to be overcome and optimized.
    \textbf{Research-only weight release.} Pre-trained weights will initially be released under an academic non-commercial licence; commercial use will require a separate agreement that enforces compliance with a no-surveillance clause.    
    \textbf{Transparent data lineage.} All training datasets are listed with licences and provenance (\autoref{sec:Dataset Introduction}); no proprietary or private images were used, reducing privacy concerns at source.